\documentclass{article}

\usepackage{xcolor}         
\definecolor{linkblue}{RGB}{0,50,120}
\usepackage[nonatbib, main, final]{neurips_2026}
\usepackage[numbers]{natbib}

\usepackage[colorlinks=true, allcolors=linkblue]{hyperref} 

\usepackage[utf8]{inputenc} 
\usepackage[T1]{fontenc}    
\usepackage{hyperref}       
\usepackage{url}            
\usepackage{booktabs}       
\usepackage{amsfonts}       
\usepackage{nicefrac}       
\usepackage{microtype}      

\usepackage{amsmath}
\usepackage{amssymb}
\usepackage{amsthm}
\usepackage{graphicx}
\usepackage{multirow}
\usepackage{subcaption}
\usepackage{caption}
\usepackage{pifont}
\usepackage{comment}
\usepackage{float}

\usepackage{algorithm}
\usepackage{algpseudocode}

\newtheorem{theorem}{Theorem}[section]

\newtheorem{definition}[theorem]{Definition}

\title{Robust and Efficient Continual Model Merging via Global Singular Subspace Separation and Restoration}

\author{
  Juncheol Shin\textsuperscript{1}\hspace{.2in}
  Byeongho Yu\textsuperscript{2}\hspace{.2in}
  Semin Kim\textsuperscript{1}\hspace{.2in}
  Eunhyeok Park\textsuperscript{1,2}\hspace{.2in}\\
  $^{1}$Graduate School of Artificial Intelligence\\
  $^{2}$Department of Computer Science \& Engineering\\
  POSTECH, South Korea\\
  \texttt{\{jchshin,bhyu418,smkim25,eh.park\}@postech.ac.kr}
}

\begin{document}

\maketitle

\begin{abstract}
Continual model merging incrementally integrates task-specific models without revisiting past models or data. However, repeated merging entangles past task components and distorts their singular geometry, making balanced task-wise allocation increasingly difficult. We propose \textbf{Global Singular Subspace Separation and Restoration (GS3R)}, which augments the merged update with a compact restoration state. At each merge, GS3R restores the retained singular geometry, separates task-wise subspaces, and globally reallocates the available rank across tasks fairly. To reduce merging-time memory overhead, we further introduce \textbf{Pre-merge Restoration and Flushing (PRF)}. Experiments on vision, NLP, and multimodal benchmarks show that GS3R provides a strong accuracy--efficiency trade-off.
\end{abstract}    

\section{Introduction}
Recent advances in machine learning have enabled effective personalization of foundation models via task-specific fine-tuning, leading to the widespread availability of specialized models across domains. In this context, \emph{model merging} has emerged as an attractive paradigm for constructing a single multi-task model from multiple task-specific models, enabling efficient reuse without additional retraining. In particular, prior methods based on singular value decomposition (SVD) or auxiliary optimization~\citep{core, knot, TSV, ISO, WUDI} achieve strong performance while incurring no inference-time overhead. However, standard model merging methods typically assume simultaneous access to all task-specific models, which is hard to satisfy in practice where models often arrive sequentially.

Recent studies have begun to address this challenge through \emph{continual model merging (CMM)}, which incrementally integrates task-specific models without retraining or storing all previous models~\citep{OPCM,DOP,NUFILT}. Despite their progress, treating this accumulated update as a single task creates two coupled problems. 
First, past task components become entangled, so the incoming task competes against all previous tasks as a single aggregate, leading to imbalanced capacity allocation. 
Second, repeated conflict-resolution operations progressively distort the singular subspaces of earlier tasks. 
Together, these effects cause \emph{task balance collapse}, where information from early tasks is increasingly underrepresented as merging proceeds.

In this work, we propose \emph{Global Singular Subspace Separation and Restoration (GS3R)}\footnote{Our code is available at \url{https://github.com/ewsn1593/GS3R}.}, a CMM framework built on a simple principle: the representation used for inference need not be the only information preserved for future merging. 
GS3R propagates the merged update together with a compact restoration state that preserves the singular geometry removed during orthogonalization. 
When a new task arrives, GS3R restores the retained pre-orthogonalization geometry, separates the recovered components into task-wise subspaces, and performs global rank allocation across all observed tasks before orthogonalizing them again for reconstruction. 
This restores task-wise access to past information without storing previous task models or explicit task-specific subspaces.

To make this process practical, we further introduce \emph{Pre-merge Restoration and Flushing (PRF)}, which restores the accumulated subspace before loading the incoming task and immediately releases the restoration state. 
GS3R therefore introduces no inference-time overhead, while its rank budget provides a controllable trade-off between additional storage and merging accuracy.

Our contributions are summarized as follows:
\begin{itemize}
    \item \textbf{Recoverable merging state.}
    GS3R preserves compact restoration information that recovers past singular geometry without storing previous task models.

    \item \textbf{Global task-wise merging.}
    Restored subspaces are separated by task, enabling balanced capacity allocation across all observed tasks.

    \item \textbf{Resource-efficient realization.}
    PRF and rank-controlled variants reduce memory and storage costs, with optional optimization providing further gains.
\end{itemize}

\section{Related Work}
\subsection{Model Merging}
Existing model merging methods can be broadly categorized into weight-space aggregation, conflict-aware task vector editing, optimization-based merging, and low-rank or SVD-based merging. A prominent line of work focuses on combining fine-tuned weights or their differences from the pre-trained model, while mitigating performance degradation caused by conflicts among multiple task updates~\citep{TIES, EMR, cat, free, TAtan, TATR, mario, adamerging}. More recently, refined optimization-based strategies~\citep{WUDI,surgery} and approaches that handle core task components via SVD have been extensively studied~\citep{core,knot,TSV,ISO,cart}. Optimization-based methods can achieve strong merging quality, but often require iterative procedures or auxiliary objectives, which increase merging-time computation. In contrast, SVD-based approaches provide structured low-rank representations of task updates and are often efficient at inference time. However, these approaches rely on the availability of complete task-specific information during merging, which is often impractical in real-world scenarios.

\subsection{Continual Model Merging}
To enable model merging in more practical settings, such as sequential adaptation to new tasks or reduced storage overhead, CMM has been introduced. 
Closely related to continual learning~\citep{overcoming, Rwalk, cont_zenke, rainbowmem, adapreplay}, CMM aims to build a unified multi-task model by incrementally incorporating new tasks. Unlike standard continual learning, where a model is updated using task data or gradients, CMM operates directly in weight space and typically assumes no access to previous task data. In CMM, task-specific models arrive sequentially, and the merger must update a unified model without revisiting all previous task models or training data. This naturally leads to a core challenge: updates from later tasks tend to overshadow those from earlier ones, causing important information to be gradually lost. To mitigate this issue, methods such as OPCM~\citep{OPCM}, DOP~\citep{DOP}, and NUFILT~\citep{NUFILT} reduce interference in sequential updates through SVD-based orthogonal projection, auxiliary optimization, null-space filtering, or their combinations.
In contrast to prior methods that primarily resolve interference locally at each step, our method explicitly maintains balance across singular components accumulated throughout the continual process, while avoiding the need to store all previous task models.

\section{Preliminaries and Challenges of Continual Model Merging}

\subsection{Problem Definition}
Consider a pre-trained model with weights $\theta_{\mathrm{pre}}$ and a set of
task-specific fine-tuned models $\{\theta_1,\ldots,\theta_N\}$ obtained from
$N$ different tasks. Standard model merging assumes that these models are
available simultaneously and seeks to construct a single model that performs
well across all tasks without jointly retraining on their original data.

\begin{definition}[Model Merging]
    \label{def:MM}
    Given task-specific models $\{\theta_1,\ldots,\theta_N\}$ and a shared
    pre-trained model $\theta_{\mathrm{pre}}$, model merging produces merged
    weights $\theta_m$ through a merge operator $\mathcal{M}$:
    \begin{equation}
        \theta_m =
        \mathcal{M}(\theta_1,\ldots,\theta_N;\theta_{\mathrm{pre}}).
    \end{equation}
\end{definition}

A common representation is the \textit{task vector}
$\tau_i=\theta_i-\theta_{\mathrm{pre}}$, which captures the parameter
displacement induced by fine-tuning. For example, Task
Arithmetic~\citep{TA} simply aggregates task vectors:
\begin{equation}
    \theta_m
    =
    \theta_{\mathrm{pre}}
    +
    \lambda \sum_{i=1}^{N}\tau_i,
\end{equation}
where $\lambda$ is a scaling coefficient. Such simple aggregation can work
surprisingly well when task updates are sufficiently compatible, but
interference between tasks often calls for more structured merging.

Several recent methods therefore operate on matrix-valued task updates in
their singular subspaces~\citep{knot,TSV,ISO}. For a task matrix $\tau_i$,
the decomposition
$\tau_i=U_i\Sigma_iV_i^\top$ represents the update as a collection of
singular components, allowing a merge operator to identify important
directions and resolve conflicts at the component level. In particular,
SVD-based approaches often concatenate singular vectors from multiple tasks
and orthogonalize or project them before reconstructing the merged
update~\citep{TSV,ISO}. When all tasks are available simultaneously, such
conflict resolution is performed once. In a continual setting, however, its
output becomes the input to subsequent merges, fundamentally changing the
problem.

Continual model merging (CMM) considers this sequential regime, where
task-specific models arrive one at a time and past models cannot be
revisited.

\begin{definition}[Continual Model Merging]
    \label{def:CMM}
    At step $t$, CMM combines the previously merged model
    $\theta_m^{t-1}$ with the newly arrived task model $\theta_t$ and
    optionally maintains a compact auxiliary state $s^{t-1}$:
    \begin{equation}
        (\theta_m^t,s^t)
        =
        \tilde{\mathcal{M}}
        (\theta_m^{t-1},s^{t-1},\theta_t;\theta_{\mathrm{pre}}),
        \qquad t\geq 2.
    \end{equation}
    Previous task-specific models and their training data are not revisited.
\end{definition}

This sequential setting fundamentally differs from standard model merging. At step $t$, the merger
has access only to the incoming task and the previously merged model, rather than the individual
models or data from tasks $1,\ldots,t-1$. Past information must therefore survive through a
representation that is repeatedly reused and modified as new tasks arrive. 

\subsection{Task Balance Collapse}\label{sec:collapse}
Over time, this can
progressively bias the merged representation toward later tasks, a phenomenon we refer to as
\textbf{Task Balance Collapse}. We identify two closely related causes: entanglement of past task
information and accumulated distortion from repeated conflict resolution.

\paragraph{Entanglement of past task information.}
In standard model merging, all task updates are available together, allowing the merge operator to
compare and allocate capacity across tasks directly. In CMM, however, tasks $1,\ldots,t-1$ are
compressed into the previous merged update $\tau_m^{t-1}$, where their individual identities and
singular components are no longer directly accessible. The next merge must therefore treat this
entangled representation as a single operand against the newly arrived task. As the number of tasks
grows, this mismatch between the objects being merged and the underlying task structure makes
balanced, task-wise capacity allocation increasingly difficult.

\paragraph{Accumulated distortion from repeated conflict resolution.}
This loss of task-wise access is further compounded by the transformations used to reduce
interference. SVD-based merging commonly orthogonalizes or projects singular subspaces before
reconstructing the merged update. While standard merging applies such conflict resolution once to
the full set of tasks, CMM repeatedly applies it to a representation that has already been modified by
previous merges. Consequently, directions inherited from earlier tasks undergo more transformations
and gradually drift from their original geometry. At the same time, joint orthogonalization obscures
the geometric relations that distinguish singular components from different tasks. Thus, the
representation becomes not only increasingly distorted, but also harder to disentangle at the next
merge.

Together, these effects reveal a central requirement of CMM: the merger must preserve not only an
effective model for the current tasks, but also enough structure to recover the task-wise geometry
needed by future merges.
\begin{figure}[t]
     \centering
         \includegraphics[width=1.0\columnwidth]{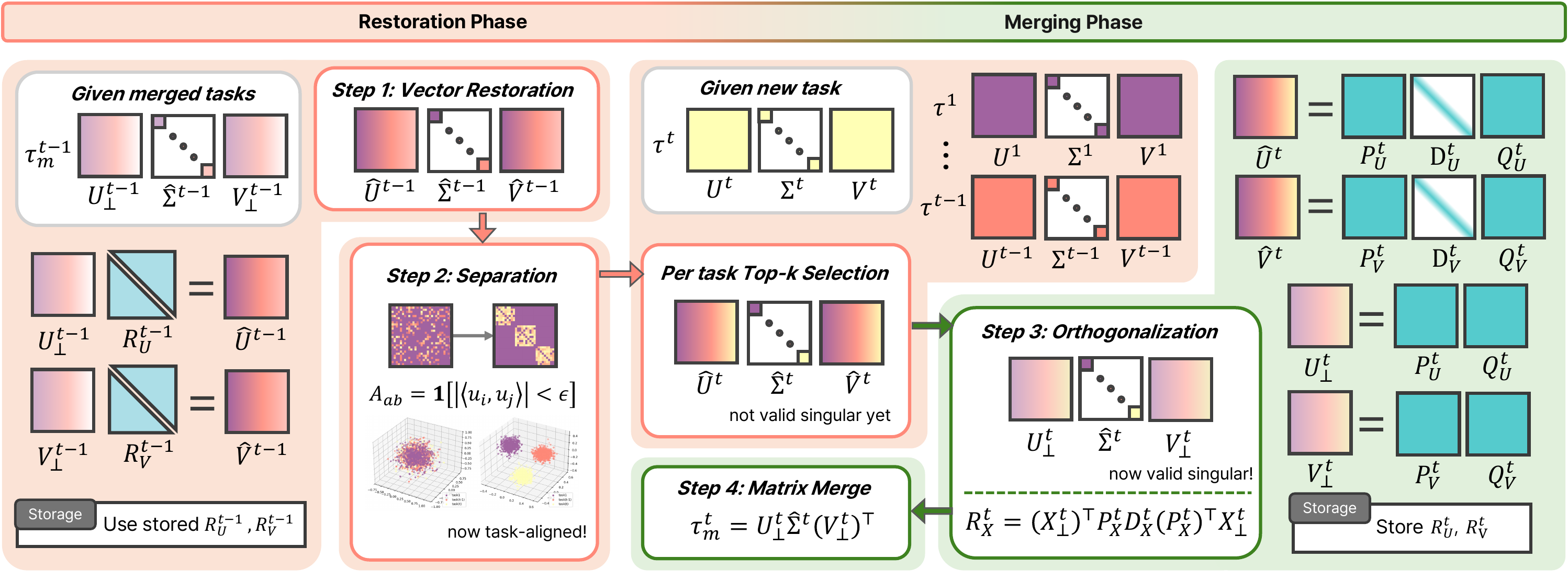}
     \caption{Overall process of GS3R. Each loop restores the task-wise geometry from the previous state, performs global task-wise merging, and stores restoration metadata for the next step.}
     \label{fig:overall}
\end{figure}

\section{Proposed Method: GS3R}
\label{sec:method}

Building on this requirement, we propose \textbf{Global Singular Subspace Separation and Restoration (GS3R)}, which reformulates continual SVD-based merging by changing the state propagated across merge steps. Rather than retaining only the merged update, GS3R propagates a recoverable state
\begin{align}
    \mathcal{S}^t
    =
    \left(
    \tau_m^t,\,
    R_U^t,\,
    R_V^t
    \right),
    \qquad
    \tau_m^t
    =
    U_\perp^t \hat{\Sigma}^t (V_\perp^t)^\top ,
    \label{eq:gs3r_state}
\end{align}
where $\tau_m^t$ is the update used for inference, while $R_U^t$ and $R_V^t$ are compact metadata that preserve the geometry needed by the next merge. At each step, GS3R restores this geometry, recovers task-wise subspaces, allocates the rank budget across all observed tasks, and re-orthogonalizes the selected components for conflict-reduced reconstruction. The resulting update is used for inference, while the restoration state is carried forward to the next step. All operations are applied independently to each matrix-valued parameter, and we omit the layer index for clarity.

At step $t$, let $\tau^t \in \mathbb{R}^{m\times n}$ denote the incoming task matrix and $r\le d=\min(m,n)$ the rank budget. We first decompose
\begin{align}
    \tau^t = U_t \Sigma_t V_t^\top,
    \label{eq:new_task_svd}
\end{align}
exposing the incoming task as singular components to be merged with the recovered components of previous tasks.

\subsection*{Step 1: Vector Restoration}
\label{sec:vector_restoration_step}

At the beginning of step $t$, the previous merged update has the orthogonalized form
\begin{align}
    \tau_m^{t-1}
    =
    U_\perp^{t-1}
    \hat{\Sigma}^{t-1}
    (V_\perp^{t-1})^\top .
    \label{eq:previous_merged_state}
\end{align}
$U_\perp^{t-1}$ and $V_\perp^{t-1}$ are valid orthogonal bases but no longer the original concatenated singular-vector bases used before the previous orthogonalization. In Figure~\ref{fig:overall}, this corresponds to the state stored under ``Given merged tasks.'' GS3R therefore first restores the pre-orthogonalized bases using the restoration matrices stored at the previous step:
\begin{align}
    \hat{U}^{t-1}
    =
    U_\perp^{t-1} R_U^{t-1},
    \qquad
    \hat{V}^{t-1}
    =
    V_\perp^{t-1} R_V^{t-1}.
    \label{eq:restore_previous_basis}
\end{align}
This reverses the stretch removed by the previous orthogonalization, so that the columns of $\hat{U}^{t-1}$ and $\hat{V}^{t-1}$ again carry the retained singular directions in pre-orthogonalized form. Crucially, this restoration makes task-wise separation possible by recovering the orthogonality structure needed to identify which retained components originated from the same task.

\subsection*{Step 2: Global Singular Subspace Separation and Top-$k$ Selection}
\label{sec:global_selection}

With this geometry restored, GS3R next recovers the hidden task-wise structure of the accumulated update. The restored $\hat{U}^{t-1}$ and $\hat{V}^{t-1}$ contain singular directions from tasks $1,\ldots,t-1$, but with interleaved columns whose task identities are not stored explicitly. GS3R infers their task-wise grouping from a basic SVD property: singular vectors from the same task are mutually orthogonal by construction, whereas those from different tasks need not be. Mutual orthogonality therefore provides a natural cue for recovering task groups. Visualization and further analysis of inter- and intra-task orthogonality are provided in Appendix~\ref{appendix:orth}.

We build an affinity graph over the columns of $\hat{U}^{t-1}$. For two restored left singular vectors $\hat{u}_a$ and $\hat{u}_b$, we define
$A_{ab}=\mathbf{1}\left[|\langle \hat{u}_a,\hat{u}_b\rangle|<\epsilon\right]$
with a small tolerance $\epsilon$. We then apply spectral clustering~\citep{spec-clust} and determine the cluster count using the eigengap heuristic. Details and empirical validation of the spectral clustering procedure are provided in Appendix~\ref{appendix:spectral_clustering}. The cluster assignment is propagated to the paired columns of $\hat{V}^{t-1}$ and to the corresponding entries of $\hat{\Sigma}^{t-1}$, yielding task-wise components:
\begin{align}
    \left\{
        U_i,\Sigma_i,V_i
    \right\}_{i=1}^{t-1},
    \qquad
    \hat{U}^{t-1}\hat{\Sigma}^{t-1}(\hat{V}^{t-1})^\top
    =
    \sum_{i=1}^{t-1} U_i\Sigma_iV_i^\top .
    \label{eq:taskwise_recovered_components}
\end{align}
The next merge therefore operates on task-wise singular subspaces rather than on ``one previous merged task plus one new task,'' removing the granularity mismatch identified in Section~\ref{sec:collapse}.

Combined with Eq.~(\ref{eq:new_task_svd}), we now have disentangled singular components for all tasks $i=1,\ldots,t$. GS3R then performs global, task-balanced selection under the rank budget $r$. With $k=\left\lfloor r/t \right\rfloor$ and $\rho=r-tk$, we take the top-$k$ components from every task and assign the remaining $\rho$ slots to the tasks with the largest next singular values. Letting $\mathcal{I}_i$ denote the selected indices for task $i$, the selected components are
\begin{align}
    \hat{U}^{t}
    &=
    [\, U_1^{(\mathcal{I}_1)},\ldots,U_t^{(\mathcal{I}_t)} \,],
    \qquad
    \hat{\Sigma}^{t}
    =
    \operatorname{diag}
    \left(
    \Sigma_1^{(\mathcal{I}_1)},\ldots,
    \Sigma_t^{(\mathcal{I}_t)}
    \right),
    \nonumber\\
    \hat{V}^{t}
    &=
    [\, V_1^{(\mathcal{I}_1)},\ldots,V_t^{(\mathcal{I}_t)} \,].
    \label{eq:selected_V}
\end{align}
This corresponds to the ``Per-task Top-$k$ Selection'' block in Figure~\ref{fig:overall}. Capacity is thus allocated at the task granularity: singular values prioritize directions within each task, while task-wise separation prevents the incoming task from competing against the entire past as a single aggregate.

\subsection*{Step 3: Orthogonalization and Restoration-State Update}
\label{sec:orthogonalization}

The matrices $\hat{U}^{t}$ and $\hat{V}^{t}$ are not yet valid singular vector bases since vectors from different tasks may be correlated. Direct use of $\hat{U}^{t}\hat{\Sigma}^{t}(\hat{V}^{t})^\top$ therefore still carries cross-task interference. Following SVD-based conflict-resolution methods~\citep{TSV,ISO}, GS3R orthogonalizes the concatenated bases via compact SVDs,
\begin{align}
    \hat{U}^{t}
    =
    P_U^t D_U^t (Q_U^t)^\top,
    \qquad
    \hat{V}^{t}
    =
    P_V^t D_V^t (Q_V^t)^\top,
    \label{eq:basis_svd}
\end{align}
and replaces them with their polar factors:
\begin{align}
    U_\perp^t
    =
    P_U^t (Q_U^t)^\top,
    \qquad
    V_\perp^t
    =
    P_V^t (Q_V^t)^\top .
    \label{eq:orthogonalized_basis}
\end{align}
The polar factor is the closest orthogonal basis to the original concatenation in Frobenius norm. Thus, Eq.~(\ref{eq:orthogonalized_basis}) can be interpreted as the least-distorting orthogonalization of the selected task subspaces, resolving cross-task correlations while keeping the selected singular values $\hat{\Sigma}^t$ unchanged.

However, the same orthogonalization that resolves cross-task interference would recreate the accumulated distortion identified in Section~\ref{sec:collapse} if the removed geometry were discarded. GS3R therefore retains the discarded stretch as compact restoration matrices. For $X\in\{U,V\}$,
\begin{align}
    R_X^t
    =
    Q_X^t D_X^t (Q_X^t)^\top.
    \label{eq:right_restoration_matrix}
\end{align}
A direct substitution gives
$\hat{X}^{t}=X_\perp^tR_X^t$,
so $R_U^t$ and $R_V^t$ make the pre-orthogonalized bases exactly recoverable at the next merge. The equivalent construction used in Figure~\ref{fig:overall} and the detailed proof are provided in Appendix~\ref{appendix:proof}.

The restoration matrices are compact. If GS3R keeps rank $r$, then $R_U^t,R_V^t\in\mathbb{R}^{r\times r}$. They are symmetric positive semidefinite, requiring only the upper triangular entries to be stored. The resulting storage overhead is $\mathcal{O}(r^2)$ per layer, independent of the number of merged tasks. Importantly, these matrices are used only during merging and introduce no inference-time overhead.

\subsection*{Step 4: Merged Matrix Construction}
\label{sec:merged_matrix_construction}

Finally, GS3R constructs the merged update at step $t$ using the orthogonalized bases and selected singular values:
\begin{align}
    \tau_m^t
    =
    U_\perp^t
    \hat{\Sigma}^t
    (V_\perp^t)^\top.
    \label{eq:merged_matrix}
\end{align}
Together with the newly computed restoration matrices, this returns the algorithm to the state in Eq.~(\ref{eq:gs3r_state}). The merged update $\tau_m^t$ is used for inference, while $(R_U^t,R_V^t)$ are retained for the next merge. This closes the continual loop:
\textit{restore $\rightarrow$ separate $\rightarrow$ select $\rightarrow$ orthogonalize $\rightarrow$ merge and store}.

\subsection{Pre-merge Restoration and Flushing}
\label{sec:prf}

The restoration matrices are compact, but peak memory can still become a practical bottleneck if the system loads the previous merged model, the new task model, and the restoration matrices simultaneously. We therefore introduce \textbf{Pre-merge Restoration and Flushing (PRF)}, an implementation strategy that changes only the execution order, not the mathematical output.

PRF exploits the fact that $R_U^{t-1}$ and $R_V^{t-1}$ are needed only for Eq.~(\ref{eq:restore_previous_basis}). When a merge request arrives, GS3R first loads the previous merged model and the restoration matrices, restores $\hat{U}^{t-1}$ and $\hat{V}^{t-1}$, and immediately flushes the restoration matrices from memory. Only after this restoration phase does it load the new task model. The restoration matrices and new task weights therefore never need to coexist in memory, preserving the benefit of restoration while limiting the peak memory overhead.

\subsection{Resource-Aware Variants}
\label{sec:variants}

\paragraph{Truncated GS3R.}
The rank budget $r$ controls the storage--accuracy trade-off. Full GS3R sets $r=d$, while truncated GS3R uses $r<d$ and retains only the most informative singular directions. This reduces additional storage from $\mathcal{O}(d^2)$ to $\mathcal{O}(r^2)$ per layer, while leaving the rest of the pipeline unchanged within the retained rank-$r$ subspace.

\paragraph{Auxiliary optimization.}
GS3R is data-free and optimization-free by default, but can also serve as a structured initialization for optimization-based merging when extra compute is available. Since Step~2 recovers task-wise components in factorized form $\tau_i=U_i\Sigma_iV_i^\top$, we initialize the optional optimization from the pre-orthogonalized reconstruction
$\tilde{\tau}^t=\hat{U}^t\hat{\Sigma}^t(\hat{V}^t)^\top$,
which retains the recovered task geometry, rather than Eq.~(\ref{eq:merged_matrix}), and minimize the auxiliary objective from~\cite{WUDI}:
\begin{align}
    \mathcal{L}(\tau)
    =
    \sum_i
    \frac{
        \left\|
        \left(
            \tau - U_i\Sigma_iV_i^\top
        \right)
        \left(
            U_i\Sigma_iV_i^\top
        \right)^\top
        \right\|_F^2
    }{
        \left\|
            U_i\Sigma_iV_i^\top
        \right\|_F^2
    } .
    \label{eq:auxiliary_objective}
\end{align}
Letting the optimized update be $\tau_{\mathrm{opt}}^t$, we store the residual
\begin{align}
    \Delta^t
    =
    \tau_{\mathrm{opt}}^t - \tau_m^t,
    \label{eq:optimization_residual}
\end{align}
to preserve the GS3R loop invariant. Before the next merge, GS3R subtracts $\Delta^t$ to recover $\tau_m^t$ in Eq.~(\ref{eq:merged_matrix}) and then applies restoration as usual. Auxiliary optimization therefore remains a modular add-on without changing the core GS3R pipeline.

\begin{table}[t]
  \centering
  \caption{Comparison of continual model merging performance for ViT models on vision tasks. We report the mean and standard deviation of accuracy (ACC) and backward transfer (BWT). STO denotes the relative storage overhead against the merged model. The horizontal line separates optimization-free (top) and optimization-based (bottom) methods. Best results within each group are indicated in bold, and second-best results are underlined. “C.” stands for Continual. }
  \label{tab:vit}
  \small
  \setlength{\tabcolsep}{4pt}
  \scalebox{0.78}{
  \begin{tabular}{cc|ccc|ccc|ccc}
    \toprule
                                                             & \multirow{3}{*}{\textbf{Method}} & \multicolumn{3}{c}{ViT-B/32} & \multicolumn{3}{c}{ViT-B/16} & \multicolumn{3}{c}{ViT-L/14}                                                                                                                                                 \\
    \cmidrule(lr){3-5} \cmidrule(lr){6-8} \cmidrule(lr){9-11}
                                                             &                         & 8 tasks                      & 14 tasks                     & 20 tasks                     & 8 tasks               & 14 tasks              & 20 tasks              & 8 tasks               & 14 tasks              & 20 tasks              \\
    \midrule
                                                             & Pre-Trained             & 48.1                         & 56.9                         & 55.6                         & 55.4                  & 62.0                  & 59.8                  & 64.9                  & 69.1                  & 65.6                  \\
                                                             & Fine-Tuned              & 90.4                         & 89.3                         & 89.8                         & 92.4                  & 91.3                  & 91.6                  & 94.3                  & 93.4                  & 93.5                  \\
                                                             & C. Fine-Tuned           & 79.8                         & 67.4                         & 62.6                         & 82.9                  & 72.2                  & 68.2                  & 90.0                  & 70.9                  & 77.7                  \\
    \midrule
    \multirow{9}{*}{\rotatebox[origin=c]{90}{ACC$\uparrow$}} & Average (SWA)           & 66.3\scriptsize$\pm$0.0                 & 65.4\scriptsize$\pm$0.0                 & 61.1\scriptsize$\pm$0.0                 & 72.3\scriptsize$\pm$0.0          & 69.7\scriptsize$\pm$0.0          & 64.8\scriptsize$\pm$0.0          & 80.0\scriptsize$\pm$0.0          & 77.5\scriptsize$\pm$0.0          & 71.1\scriptsize$\pm$0.0          \\
                                                             & C. Task Arithmetic      & 67.5\scriptsize$\pm$0.0                 & 66.5\scriptsize$\pm$0.0                 & 60.6\scriptsize$\pm$0.0                 & 77.1\scriptsize$\pm$0.0          & 70.9\scriptsize$\pm$0.0          & 64.2\scriptsize$\pm$0.0          & 82.1\scriptsize$\pm$0.0          & 77.9\scriptsize$\pm$0.0          & 70.3\scriptsize$\pm$0.0          \\
                                                             & C. Ties-Merging         & 49.0\scriptsize$\pm$10.2                & 66.2\scriptsize$\pm$0.6                 & 59.9\scriptsize$\pm$0.7                 & 66.8\scriptsize$\pm$3.7          & 70.5\scriptsize$\pm$0.8          & 63.0\scriptsize$\pm$1.6          & 64.3\scriptsize$\pm$7.0          & 78.0\scriptsize$\pm$0.6          & 68.3\scriptsize$\pm$0.9          \\
                                                             & OPCM    & 75.5\scriptsize$\pm$0.5        & 71.9\scriptsize$\pm$0.3        & 65.7\scriptsize$\pm$0.2        & 81.8\scriptsize$\pm$0.3 & 77.1\scriptsize$\pm$0.5 & 70.3\scriptsize$\pm$0.2 & 87.0\scriptsize$\pm$0.4 & 83.5\scriptsize$\pm$0.2 & 76.0\scriptsize$\pm$0.2 \\
                                                             & GS3R ($r=256$) & \underline{81.8}\scriptsize$\pm$0.2        & \underline{76.5}\scriptsize$\pm$0.3                 & \underline{70.9}\scriptsize$\pm$0.5                 & \underline{86.7}\scriptsize$\pm$0.1          & \underline{80.3}\scriptsize$\pm$0.2          & \underline{76.1}\scriptsize$\pm$0.5          & \underline{89.4}\scriptsize$\pm$0.4 & \underline{86.4}\scriptsize$\pm$0.1          & \underline{81.7}\scriptsize$\pm$0.3          \\
                                                             & GS3R    & \textbf{82.7}\scriptsize$\pm$0.2        & \textbf{78.7}\scriptsize$\pm$0.3        & \textbf{72.7}\scriptsize$\pm$0.6        & \textbf{86.9}\scriptsize$\pm$0.1 & \textbf{81.8}\scriptsize$\pm$0.3 & \textbf{77.7}\scriptsize$\pm$0.5 & \textbf{90.3}\scriptsize$\pm$0.5 & \textbf{88.0}\scriptsize$\pm$0.3 & \textbf{82.7}\scriptsize$\pm$0.7 \\
                                                             \cmidrule(lr){2-11}
                                                             & DOP    & 78.3\scriptsize$\pm$1.6        & 73.0\scriptsize$\pm$1.2        & 68.5\scriptsize$\pm$1.6        & 84.5\scriptsize$\pm$0.8 & 76.2\scriptsize$\pm$1.2 & 72.2\scriptsize$\pm$1.8 & 88.3\scriptsize$\pm$0.7 & 85.5\scriptsize$\pm$0.8 & 80.7\scriptsize$\pm$0.9 \\
                                                             & NUFILT    & \underline{83.6}\scriptsize$\pm$0.2        & \underline{78.0}\scriptsize$\pm$0.2        & \underline{71.0}\scriptsize$\pm$0.9        & \underline{87.3}\scriptsize$\pm$0.1 & \underline{83.1}\scriptsize$\pm$0.3 & \underline{78.1}\scriptsize$\pm$0.9 & \underline{91.6}\scriptsize$\pm$0.1 & \underline{89.2}\scriptsize$\pm$0.1 & \underline{84.7}\scriptsize$\pm$0.8 \\
                                                             
                                                             & GS3R + Opt. & \textbf{84.7}\scriptsize$\pm$0.3        & \textbf{80.0}\scriptsize$\pm$0.4                 & \textbf{73.2}\scriptsize$\pm$1.4                 & \textbf{89.2}\scriptsize$\pm$0.1          & \textbf{83.8}\scriptsize$\pm$0.3          & \textbf{79.4}\scriptsize$\pm$0.7          & \textbf{92.2}\scriptsize$\pm$0.3           &  \textbf{89.8}\scriptsize$\pm$0.3          & \textbf{85.2}\scriptsize$\pm$1.5          \\
    \midrule
    \multirow{9}{*}{\rotatebox[origin=c]{90}{BWT$\uparrow$}} & Average (SWA)           & -11.5\scriptsize$\pm$2.2                & -8.0\scriptsize$\pm$1.3                 & -7.1\scriptsize$\pm$2.1                 & -9.7\scriptsize$\pm$1.5          & -7.1\scriptsize$\pm$1.4          & -7.3\scriptsize$\pm$1.7          & -7.3\scriptsize$\pm$1.4          & -5.8\scriptsize$\pm$1.0          & -6.4\scriptsize$\pm$1.5          \\
                                                             & C. Task Arithmetic      & -9.6\scriptsize$\pm$1.5                 & \underline{-1.3}\scriptsize$\pm$0.3                 & \underline{-3.4}\scriptsize$\pm$0.4                 & -4.2\scriptsize$\pm$1.0 & \underline{-1.3}\scriptsize$\pm$0.4          & \underline{-3.6}\scriptsize$\pm$0.4          & -7.1\scriptsize$\pm$0.8          & \underline{-1.8}\scriptsize$\pm$0.3          & \underline{-3.3}\scriptsize$\pm$0.3          \\
                                                             & C. Ties-Merging         & -15.3\scriptsize$\pm$8.0                & \textbf{1.9}\scriptsize$\pm$0.6         & \textbf{-1.5}\scriptsize$\pm$0.7        & -5.5\scriptsize$\pm$0.4          & \textbf{1.4}\scriptsize$\pm$0.7  & \textbf{-1.5}\scriptsize$\pm$1.2 & -13.0\scriptsize$\pm$5.7         & \textbf{ 1.1}\scriptsize$\pm$0.4  & \textbf{-2.9}\scriptsize$\pm$1.0 \\
                                                             & OPCM    & -6.3\scriptsize$\pm$1.1        & -6.0\scriptsize$\pm$1.0                 & -7.8\scriptsize$\pm$1.5                 & -4.8\scriptsize$\pm$0.7          & -5.1\scriptsize$\pm$1.4          & -6.3\scriptsize$\pm$2.2          & \underline{-2.6}\scriptsize$\pm$1.0 & -4.3\scriptsize$\pm$0.7          & -6.5\scriptsize$\pm$1.8          \\
                                                             & GS3R ($r=256$)&    \underline{-4.9}\scriptsize$\pm$1.2   &     -6.4\scriptsize$\pm$1.2            &    -9.3\scriptsize$\pm$1.8            &  \underline{-3.1}\scriptsize$\pm$0.6       &   -5.7\scriptsize$\pm$1.1    &   -7.5\scriptsize$\pm$1.3     & -2.7\scriptsize$\pm$0.9 &  -3.5\scriptsize$\pm$0.8         &    -6.1\scriptsize$\pm$1.0       \\
                                                             & GS3R    &  \textbf{-4.3}\scriptsize$\pm$0.9      &     -5.2\scriptsize$\pm$0.9             &    -8.8\scriptsize$\pm$1.7             &   \textbf{-3.0}\scriptsize$\pm$0.5        &   -4.6\scriptsize$\pm$1.1      &    -6.9\scriptsize$\pm$1.4       & \textbf{-1.9}\scriptsize$\pm$1.0 &    -2.7\scriptsize$\pm$0.8      &    -6.3\scriptsize$\pm$1.5      \\
                                                             
                                                             \cmidrule(lr){2-11}
                                                             & DOP    & -6.1\scriptsize$\pm$2.3        & \underline{-8.8\scriptsize$\pm$2.1}                 & -12.5\scriptsize$\pm$2.8                 & -3.9\scriptsize$\pm$1.0          & -7.5\scriptsize$\pm$2.2          & -11.4\scriptsize$\pm$2.8          & -3.2\scriptsize$\pm$1.2 & -4.6\scriptsize$\pm$1.0          & -8.4\scriptsize$\pm$1.4          \\
                                                             & NUFILT    & \textbf{-2.7}\scriptsize$\pm$0.7        & -5.7\scriptsize$\pm$0.9                 & \textbf{-8.9}\scriptsize$\pm$2.3                 & \textbf{-1.6}\scriptsize$\pm$0.5          & \textbf{-3.5}\scriptsize$\pm$0.6          & \textbf{-7.1}\scriptsize$\pm$1.9         & \textbf{-1.1}\scriptsize$\pm$0.3 & \textbf{-2.0}\scriptsize$\pm$0.3          & \textbf{-4.6}\scriptsize$\pm$0.7          \\
                                                             
                                                             & GS3R + Opt.&    \underline{-3.4}\scriptsize$\pm$0.8     &      \textbf{-5.0}\scriptsize$\pm$1.0          &       \underline{-10.3}\scriptsize$\pm$2.0        &    \underline{-2.0}\scriptsize$\pm$0.3      &    \underline{-4.1}\scriptsize$\pm$0.7     &    \underline{-7.6}\scriptsize$\pm$1.2     &   \underline{-1.2}\scriptsize$\pm$0.5       &   \underline{-2.1}\scriptsize$\pm$0.6 &  \underline{-7.3}\scriptsize$\pm$1.2   \\

    \midrule
     \multirow{3}{*}{\rotatebox[origin=c]{90}{STO$\downarrow$}} 
     & GS3R ($r=256$)           &             \multicolumn{3}{c|}{ 1.05$\times$ (+ 19.0 MB) }      &  \multicolumn{3}{c|}{1.06$\times$ (+ 19.1 MB)}   & \multicolumn{3}{c}{ 1.03$\times$ (+ 38.2 MB) }                   \\    
     & GS3R         &      \multicolumn{3}{c|}{1.49$\times$ (+ 170.1 MB)  }             &      \multicolumn{3}{c|}{1.50$\times$ (+ 170.2 MB)  }       &        \multicolumn{3}{c}{1.50$\times$ (+ 604.8 MB)  }              \\   
     & GS3R + Opt.           &             \multicolumn{3}{c|}{ 2.46$\times$ (+ 510.0 MB) }      &  \multicolumn{3}{c|}{2.49$\times$ (+  510.6 MB)}   & \multicolumn{3}{c}{ 2.50$\times$ (+ 1813.8 MB) }                   \\   
    \bottomrule

  \end{tabular}
  }

\end{table}

\section{Experimental Results}
\label{sec:exp}

We evaluate GS3R from three perspectives: whether recovering task-wise singular structure improves continual merging performance, how the rank-controlled restoration state trades accuracy for storage and merging cost, and how each component contributes to the overall framework. We consider vision, NLP, and multimodal benchmarks. For vision, we use CLIP~\citep{clip}-based ViT~\citep{vit} models fine-tuned on 20 downstream tasks and, following prior work~\citep{OPCM,DOP}, evaluate continual merging over suites of 8, 14, and 20 tasks. For NLP, we use Flan-T5-Base~\citep{flan} and continually merge models fine-tuned on 8 GLUE tasks~\citep{glue}. To assess scalability and generality beyond these settings, we further merge LLaVA-1.5-7B~\citep{llava} models on four multimodal benchmarks: IconQA~\citep{iconqa}, ImageNet captioning~\citep{imagenet}, ScienceQA~\citep{scienceqa}, and VizWiz-Caption~\citep{vizwiz}, which span diagram reasoning, natural image captioning, scientific question answering, and real-world low-vision scenarios.

We compare against Simple Weight Averaging (SWA)~\citep{soup}, Continual Task Arithmetic~\citep{TA}, Continual TIES-Merging~\citep{TIES}, and three recent CMM methods: OPCM~\citep{OPCM}, DOP~\citep{DOP}, and NUFILT~\citep{NUFILT}. All experiments use publicly available checkpoints~\citep{FusionBench,RobustMerge}. Results are averaged over 10 random task orders for vision and NLP benchmarks and over 6 task orders (alphabetical and 5 random) for multimodal benchmarks. Vision and NLP experiments run on an NVIDIA RTX 3090 GPU, and multimodal experiments on an NVIDIA A100-SXM4-80GB GPU. Further details on experimental settings are provided in Appendix~\ref{appendix:exp}.

\subsection{Continual Model Merging on Vision Tasks}
\label{sec:exp_vision}
For vision tasks, we evaluate GS3R under three settings: full-rank GS3R, truncated GS3R with $r=256$, and full-rank GS3R with auxiliary optimization. Table~\ref{tab:vit} summarizes the results across three ViT backbones and task sequences of increasing length. 

Without auxiliary optimization, GS3R consistently outperforms OPCM and DOP across all backbone and task-count settings. Its advantage is particularly visible on ViT-B/32 as the continual sequence grows: GS3R improves over NUFILT from 78.0 to 78.7 on 14 tasks and from 71.0 to 72.7 on 20 tasks, consistent with its goal of preserving balanced task-wise capacity as more tasks accumulate. Truncating the retained subspace to $r=256$ preserves most of this performance while substantially reducing the auxiliary storage. When additional computation is available, auxiliary optimization further improves GS3R and achieves the highest accuracy across all nine vision settings. These results show that the recovered task-wise subspaces provide a strong optimization-free merging state while remaining compatible with further refinement.

\begin{table}[t]
  \centering
  \caption{Continual model merging results for the Flan-T5-Base model on GLUE tasks. We report the mean and standard deviation of accuracy and backward transfer.}
  \label{tab:flan}
  \small
  \setlength{\tabcolsep}{4pt}
  \scalebox{0.85}{
      \begin{tabular}{ccccccccc|ccc}
        \toprule
            Method & CoLA & MNLI & MRPC & QNLI & QQP & RTE & SST2 &STSB & ACC($\uparrow$) & BWT($\uparrow$) & STO($\downarrow$) \\
        \midrule
            Pre-Trained & 69.1 & 56.5 & 76.2 & 88.4 & 82.1 & 80.1 & 91.2 & 62.2 & 75.7 & - & - \\
            Fine-Tuned & 75.0 & 83.4 & 87.5 & 91.5 & 85.4 & 85.9 & 93.6 & 88.7 & 86.4 & - & - \\
        \midrule
    
         OPCM    &   69.7      &      72.9          &     \underline{78.8}        &   90.2      &     83.8    &  \underline{82.2}   &  92.3  &   74.7     & 80.6\scriptsize$\pm$0.3  & -2.1\scriptsize$\pm$0.6 & 1.00$\times$    \\
         
         GS3R  &   \underline{70.5}      &       \underline{79.5}          &         \underline{78.8}         &   \underline{90.6}      &   \textbf{84.3}      &  \textbf{83.0}   &  \underline{92.9}  &    \underline{81.3}    &  \underline{82.6}\scriptsize$\pm$0.1  &  \underline{-1.9}\scriptsize$\pm$0.6 &  1.525$\times$\\ 
         GS3R ($r=64$)   &   \textbf{71.7}    &       \textbf{81.1}         &        \textbf{80.0}         &   \textbf{90.7}      &   \underline{84.2}     &  \underline{82.2}   & \textbf{93.1}  &     \textbf{86.4}   &  \textbf{83.7}\scriptsize$\pm$0.1 & \textbf{-1.2}\scriptsize$\pm$0.4 & 1.004$\times$   \\
         \cmidrule(lr){1-12}
         DOP   &   \underline{72.6}      &      80.9         &     \textbf{81.0}      &    90.4    &   \underline{83.4}      &  \underline{81.0}   & 92.9  &   86.1     &  \underline{83.6}\scriptsize$\pm$0.1 & \textbf{-1.3}\scriptsize$\pm$0.3 & 1.00$\times$   \\
         NUFILT    &    \textbf{72.7}     &      \textbf{83.0}          &       80.3           &   \textbf{90.9}     &    \textbf{84.0}     &  80.6   & \textbf{93.3}  & \textbf{86.9}       & \textbf{84.0}\scriptsize$\pm$0.2  & \underline{-1.4}\scriptsize$\pm$0.5 & 1.00$\times$   \\
         GS3R ($r=64$) + Opt.   &   72.0   &      \textbf{83.0}       &           \underline{80.5}       &    \underline{90.8}    &    \textbf{84.0}       &   \textbf{82.0}        & \underline{93.2} &   \underline{86.4}     & \textbf{84.0}\scriptsize$\pm$0.1 & -1.9\scriptsize$\pm$0.2  & 1.804$\times$  \\ 
        \bottomrule
      \end{tabular}
      }
\end{table}

\begin{figure}
    \centering
    \begin{subfigure}{0.45\textwidth}
        \includegraphics[width=\linewidth]{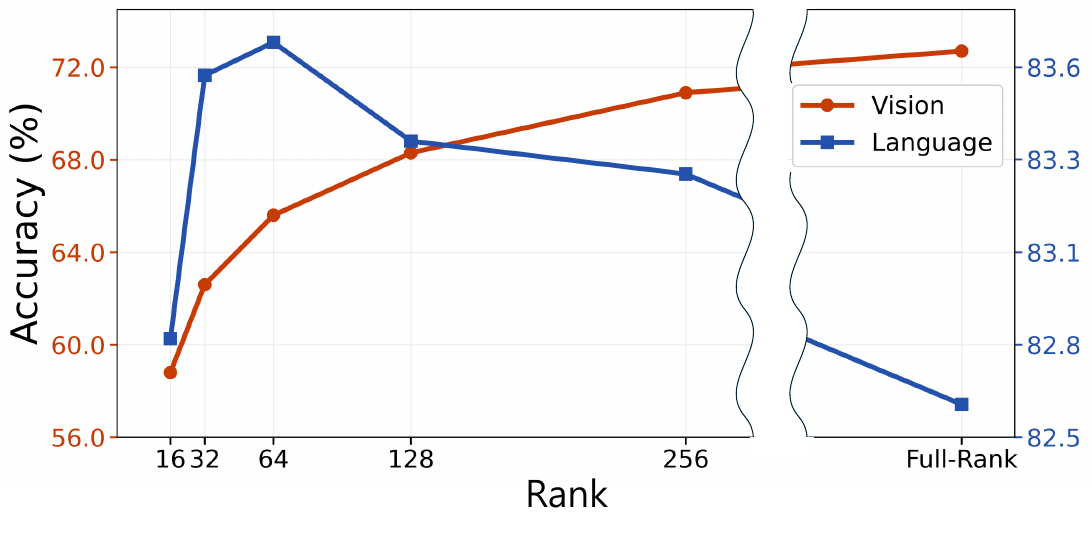}
        \caption{Accuracy on Vision and NLP tasks}
    \end{subfigure}
    \hspace{0.08\textwidth}
    \begin{subfigure}{0.40\textwidth}
        \includegraphics[width=\linewidth]{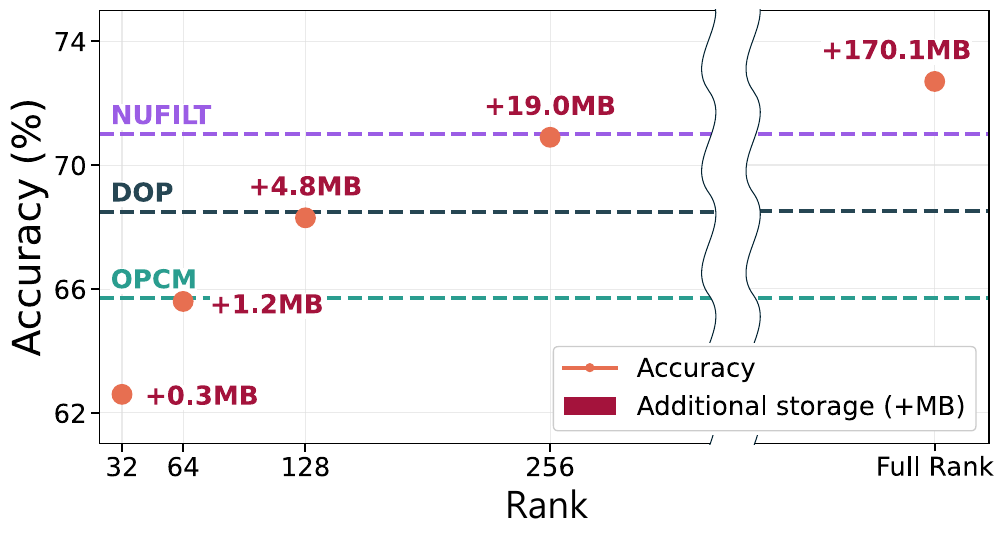}
        \caption{Storage trade-off}
    \end{subfigure}
    \caption{Accuracy and storage trade-off with varying rank $r$. (a) shows accuracy variation with varying $r$ on vision and NLP tasks. (b) shows that combining GS3R with truncated SVD yields a favorable trade-off. $r=256$ preserves strong performance while requiring only 0.05$\times$ more storage.}
    \label{fig:trade-off}

\end{figure}

\subsection{Continual Model Merging on NLP Tasks}
\label{sec:nlp_exp}

Table~\ref{tab:flan} evaluates GS3R on Flan-T5-Base continually merged over eight GLUE tasks. With a rank budget of $r=64$, GS3R reaches an average accuracy of 83.7 without any optimization and achieves the best BWT of $-1.2$ among all methods. This surpasses OPCM by 3.1 points and matches DOP while coming within 0.3 of NUFILT, even though both rely on auxiliary optimization. The low BWT is notable because it indicates that the merged model retains earlier tasks almost intact as new ones are added, which is precisely the property the restoration state is designed to preserve. Adding auxiliary optimization raises the average accuracy to 84.0, on par with NUFILT. This demonstrates that the same recoverable state formulation extends beyond vision models without a task-specific redesign.

Interestingly, the truncated variant performs better than full-rank GS3R on GLUE, unlike the monotonic trend observed on the vision benchmark. Figure~\ref{fig:trade-off}(a) shows that NLP performance peaks at an intermediate rank rather than at full rank. Our analysis in Appendix~\ref{appendix:rank_analysis} attributes this behavior to the substantially lower effective rank of several Flan-T5 task matrices. Once the informative directions are already retained, additional components contribute little to the task while still participating in joint orthogonalization, where they can increase the distortion introduced during merging. Truncation therefore serves not only as a storage mechanism, but also as an effective filter when the task updates are intrinsically low-rank.

\subsection{Continual Model Merging on Multimodal Tasks with Large-Scale Models}

Table~\ref{tab:llava} reports results for continually merging LLaVA-1.5-7B over the four multimodal tasks. As in the vision experiments, we evaluate full-rank GS3R, truncated GS3R with $r=256$, and full-rank GS3R with auxiliary optimization. Compared to the previous benchmarks, this setting involves a billion-scale backbone and open-ended generation tasks rather than classification. Nonetheless, the same recoverable-state formulation transfers without modification.

Under the original scorers, all GS3R variants outperform every baseline, with optimization-free GS3R ($r=256$) reaching 69.5 against 66.6 for NUFILT. However, part of this margin stems from IconQA, the only task scored by exact string match, where merged models often append extra tokens to the correct answer. We therefore also report ACC$^\dagger$, which scores IconQA with a format-tolerant rule for all methods and thus measures answer correctness more directly (Appendix~\ref{appendix:llava_eval}). Under this protocol, GS3R without any optimization attains 72.1, within 0.6 of NUFILT, and truncating to $r=256$ yields 72.7, matching NUFILT with only 0.2\% additional storage and a standard deviation of just 0.1 across task orders. With auxiliary optimization, GS3R reaches 73.5, surpassing NUFILT by 0.8 points and attaining the best score among optimization-based methods on all four tasks. GS3R therefore matches or surpasses NUFILT under both protocols, without any gradient computation in the optimization-free case. Overall, the recovered task-wise singular structure remains an effective and scalable merging state beyond vision and NLP, extending to large-scale multimodal models.
\begin{table}[t]
\centering
\small
\setlength{\tabcolsep}{4pt}
\caption{Continual model merging results of LLaVA-1.5-7B on multimodal tasks. ACC$^\dagger$ scores IconQA with a format-tolerant rule applied identically to all methods, with the corresponding IconQA accuracy in parentheses. Detailed explanations are provided in Appendix~\ref{appendix:llava_eval}.}
\label{tab:llava}
\begin{tabular}{ccccc|ccc}
\toprule
Method & IconQA & Image & SciQA & VizWiz & ACC($\uparrow$) & ACC$^\dagger$($\uparrow$) & STO($\downarrow$) \\
\midrule
Pre-Trained & 14.1 & 40.9 & 61.7 & 41.0 & 39.4 & -- & -- \\
Fine-Tuned  & 75.5 & 96.0 & 83.7 & 64.8 & 80.0 & -- & -- \\
\midrule
C. Task Arithmetic & 55.7 (55.9) & 75.4 & 77.2 & 46.7 & 63.8{\scriptsize$\pm$0.0} & 63.8{\scriptsize$\pm$0.0} & 1.00$\times$ \\
C. Ties-Merging    & \textbf{57.6} (63.7) & 84.1 & 79.8 & 44.6 & 66.5{\scriptsize$\pm$3.9} & 68.1{\scriptsize$\pm$1.6} & 1.00$\times$ \\
OPCM               & 54.6 (63.1) & 83.1 & \underline{80.9} & 45.6 & 66.0{\scriptsize$\pm$0.7} & 68.2{\scriptsize$\pm$0.8} & 1.00$\times$ \\
GS3R ($r{=}256$)   & 56.2 (69.3) & \textbf{93.8} & \underline{80.9} & \underline{47.0} & \textbf{69.5}{\scriptsize$\pm$0.2} & \textbf{72.7}{\scriptsize$\pm$0.1} & 1.002$\times$ \\
GS3R               & \underline{56.9} (68.2) & \underline{90.6} & \textbf{82.1} & \textbf{47.4} & \underline{69.2}{\scriptsize$\pm$0.8} & \underline{72.1}{\scriptsize$\pm$0.6} & 1.535$\times$ \\
\midrule
NUFILT             & \underline{45.3} (69.6) & \underline{94.0} & \underline{80.8} & \underline{46.3} & \underline{66.6}{\scriptsize$\pm$5.9} & \underline{72.7}{\scriptsize$\pm$1.0} & 1.00$\times$ \\
GS3R + Opt.        & \textbf{58.7} (71.2) & \textbf{94.2} & \textbf{81.7} & \textbf{46.7} & \textbf{70.3}{\scriptsize$\pm$1.1} & \textbf{73.5}{\scriptsize$\pm$1.0} & 2.455$\times$ \\
\bottomrule
\end{tabular}
\end{table}

\subsection{Storage Accuracy Trade-off} 

The rank budget $r$ directly controls the size of the recoverable state carried between merge steps. We therefore examine how aggressively this state can be truncated while preserving merging quality. Figure~\ref{fig:trade-off}(b) reports the resulting accuracy--storage trade-off for ViT-B/32 over 20 vision tasks.

Full-rank GS3R requires 170.1 MB of additional storage and achieves 72.7\% accuracy. Reducing the budget to $r=256$ lowers the additional storage to only 19.0 MB, an $8.9\times$ reduction, while retaining 70.9\% accuracy, a drop of 1.8 points. Thus, most of the benefit of the recoverable state can be retained with only a small fraction of its full-rank storage. This makes the rank budget a direct and controllable mechanism for trading merging accuracy against deployment storage.

\subsection{Merging Overhead Comparison}

A useful continual merger must also keep the cost of each update practical. Table~\ref{tab:run} therefore compares the final accuracy after merging 20 tasks on ViT-B/32 together with the peak GPU memory and elapsed time of a single merge, measured on an NVIDIA RTX 3090 GPU. With PRF, GS3R reaches 72.7\% accuracy using 180.0 MB of peak GPU memory and 17.6 seconds per merge. This is only a modest increase over the lightweight OPCM baseline, while providing a 7.0 point accuracy gain.

Compared with the stronger NUFILT baseline, GS3R improves accuracy from 71.0\% to 72.7\%, while reducing peak GPU memory from 2449.3 MB to 180.0 MB and merge time from 26.8 to 17.6 seconds. These results show that the additional restoration state does not translate into a large runtime or memory burden when combined with PRF. Auxiliary optimization can further raise the accuracy to 73.2\%, at the expected cost of higher memory consumption and longer merging time, providing an additional accuracy--compute operating point when resources permit.

\begin{minipage}[t]{0.51\linewidth}
\vspace{0pt} 
      \centering
      \small
      \resizebox{1\columnwidth}{!}{
      \begin{tabular}{c|ccc}
        \toprule
            Method   & Accuracy ($\%$) & GPU Mem. (MB) &  Time (s)   \\
        \midrule 
            OPCM & 65.7\scriptsize$\pm$0.2 & 134.2 & 14.6  \\
            DOP & 68.5\scriptsize$\pm$1.6 & 259.4 & 100.8 \\
            NUFILT & 71.0\scriptsize$\pm$0.9 & 2449.3 & 26.8 \\
            GS3R & 72.7\scriptsize$\pm$0.6 & 180.0 & 17.6 \\
            GS3R + Opt. & 73.2\scriptsize$\pm$1.4 & 584.1 & 60.6\\
        \bottomrule
      \end{tabular}
      }
       \vspace{0pt} 
      \captionof{table}{Merging overhead comparison on ViT-B/32, measured on an NVIDIA RTX 3090 GPU. Our method achieves high accuracy with low latency and minimal GPU memory overhead.}    
      \label{tab:run}
\end{minipage}
\hfill
\begin{minipage}[t]{0.48\linewidth}
\vspace{0pt}
  \centering
  \resizebox{1\columnwidth}{!}{
  \begin{tabular}{cccc ccc|c}
    \toprule
    \multirow{2}{*}{\textbf{G}} & \multirow{2}{*}{\textbf{R}} & \multirow{2}{*}{\textbf{T}} & \multirow{2}{*}{\textbf{O}} &
    \multicolumn{4}{c}{\textbf{ViT-B/32}} \\
    \cmidrule{5-8}
     &  &  &  & 8 tasks & 14 tasks & 20 tasks & + Storage \\
    \midrule
    \textcolor{red}{\ding{55}} & \textcolor{red}{\ding{55}} & \textcolor{red}{\ding{55}} & \textcolor{red}{\ding{55}} & 67.8 & 64.9 & 55.6 & 0 MB \\
    \textcolor[RGB]{0,150,0}{\checkmark} & \textcolor{red}{\ding{55}} & \textcolor{red}{\ding{55}} & \textcolor{red}{\ding{55}} & 67.8 & 65.0 & 55.6  & 0 MB \\
    \textcolor{red}{\ding{55}} & \textcolor[RGB]{0,150,0}{\checkmark} & \textcolor{red}{\ding{55}} & \textcolor{red}{\ding{55}} & 71.8 & 70.2 & 61.7 & 170.1 MB \\
    \midrule
    \textcolor[RGB]{0,150,0}{\checkmark} & \textcolor[RGB]{0,150,0}{\checkmark} & \textcolor{red}{\ding{55}} & \textcolor{red}{\ding{55}} & 82.7 & 78.7 & 72.7 & 170.1 MB \\
    \textcolor[RGB]{0,150,0}{\checkmark} & \textcolor[RGB]{0,150,0}{\checkmark} & \textcolor[RGB]{0,150,0}{\checkmark} & \textcolor{red}{\ding{55}} & 81.8  & 76.5 & 70.9  & 19.0 MB \\
    \textcolor[RGB]{0,150,0}{\checkmark} & \textcolor[RGB]{0,150,0}{\checkmark} & \textcolor{red}{\ding{55}} &\textcolor[RGB]{0,150,0}{\checkmark}  & 84.7  & 80.0 & 73.2  & 510.0 MB \\
    \bottomrule
  \end{tabular}
  }
  \captionof{table}{Ablation study on components of GS3R. 
  \textbf{G}: Global singular subspace separation, \textbf{R}: singular vector Restoration, \textbf{T}: Truncated SVD, \textbf{O}: auxiliary Optimization.}

  \label{tab:abl}
\end{minipage}
\subsection{Ablation Study}

We ablate the key components of GS3R on ViT-B/32 to understand how restoration and task-wise separation interact. The results are summarized in Table~\ref{tab:abl}. Applying global singular subspace separation alone provides essentially no improvement over the baseline, because the accumulated bases remain orthogonalized and no longer expose the task-wise geometry required for reliable separation. In contrast, restoration alone already yields substantial gains, improving accuracy by 4.0, 5.3, and 6.1 points on the 8-, 14-, and 20-task settings, respectively. However, without subsequent task-wise separation and balanced rank allocation, the restored components still compete as an entangled aggregate.

Combining restoration with global separation produces a much larger improvement, reaching 82.7\%, 78.7\%, and 72.7\% across the three settings. This confirms the intended dependency between the two components: restoration recovers the geometric structure needed to identify retained task components, while separation converts that structure into task-wise capacity allocation. The remaining variants characterize the resource trade-off of this core pipeline. Truncating the retained subspace to $r=256$ reduces the storage from 170.1 MB to 19.0 MB while preserving most of the accuracy, whereas auxiliary optimization further improves performance at the cost of additional storage and computation.

\begin{figure}[t]
    \centering
    \begin{subfigure}{0.2\textwidth}
        \includegraphics[width=\linewidth]{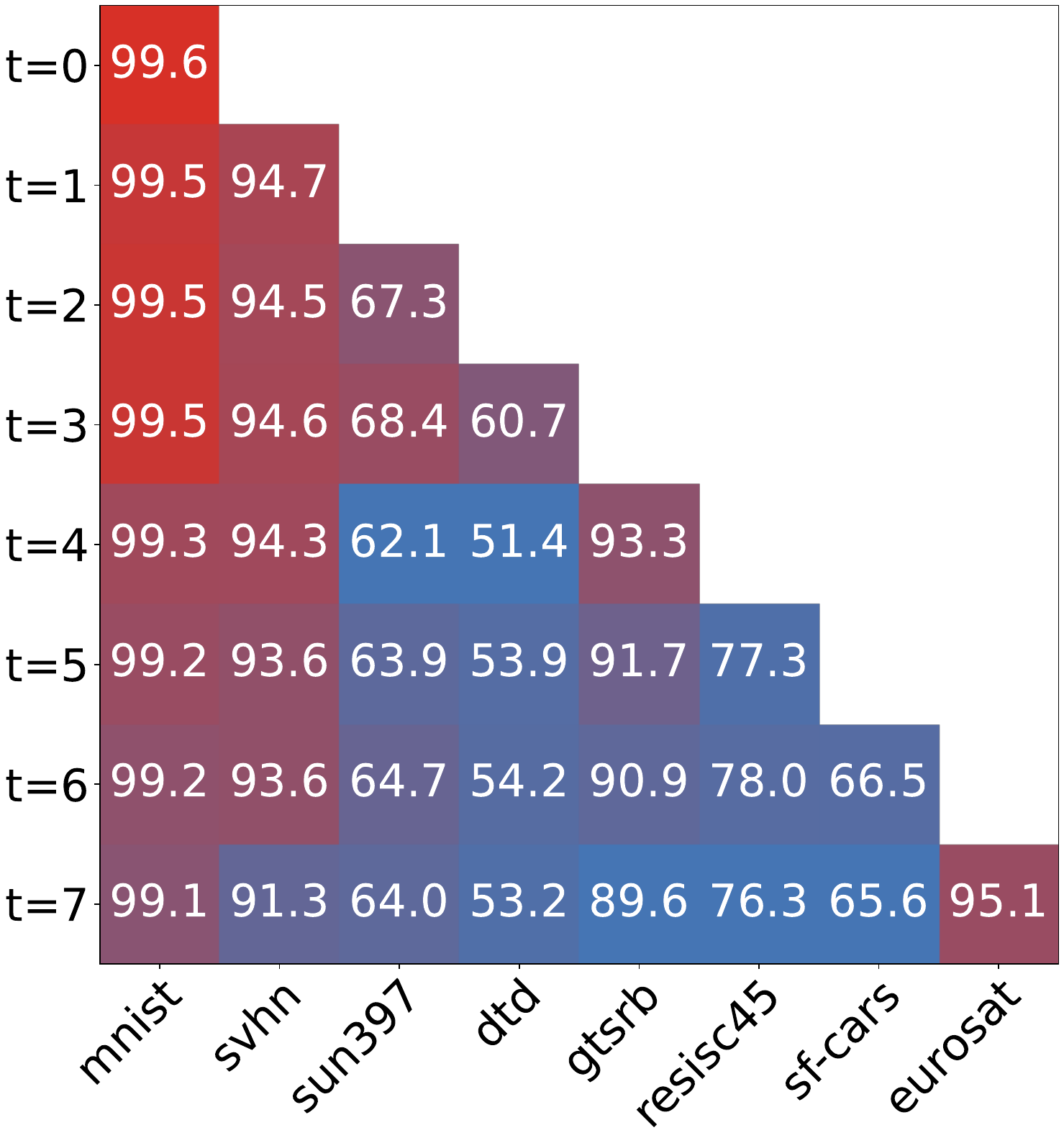}
        \caption{DOP}
    \end{subfigure}
    \hspace{20pt}
    \begin{subfigure}{0.2\textwidth}
        \includegraphics[width=\linewidth]{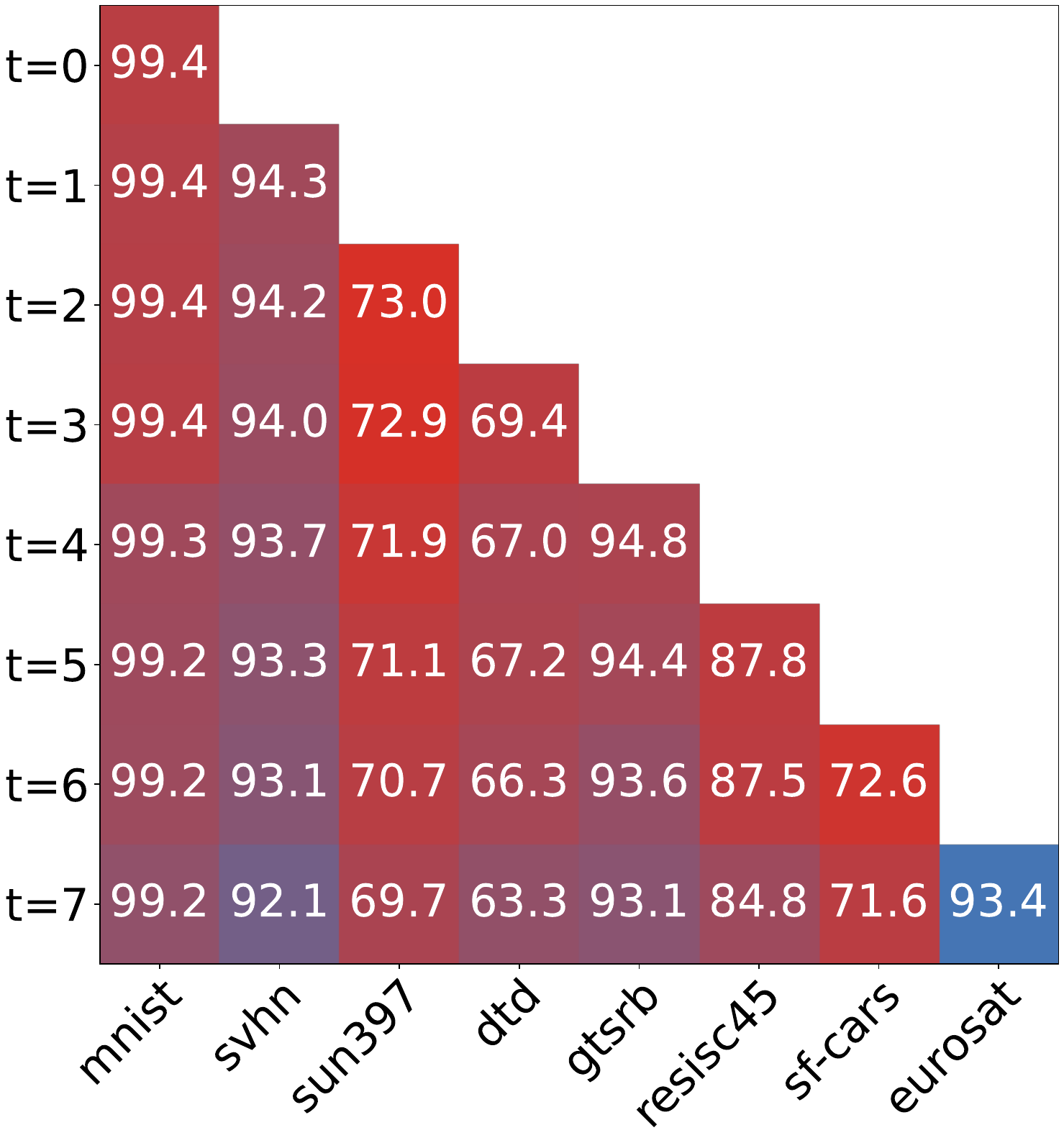}
        \caption{NUFILT}
    \end{subfigure}
    \hspace{20pt}
    \begin{subfigure}{0.2\textwidth}
        \includegraphics[width=\linewidth]{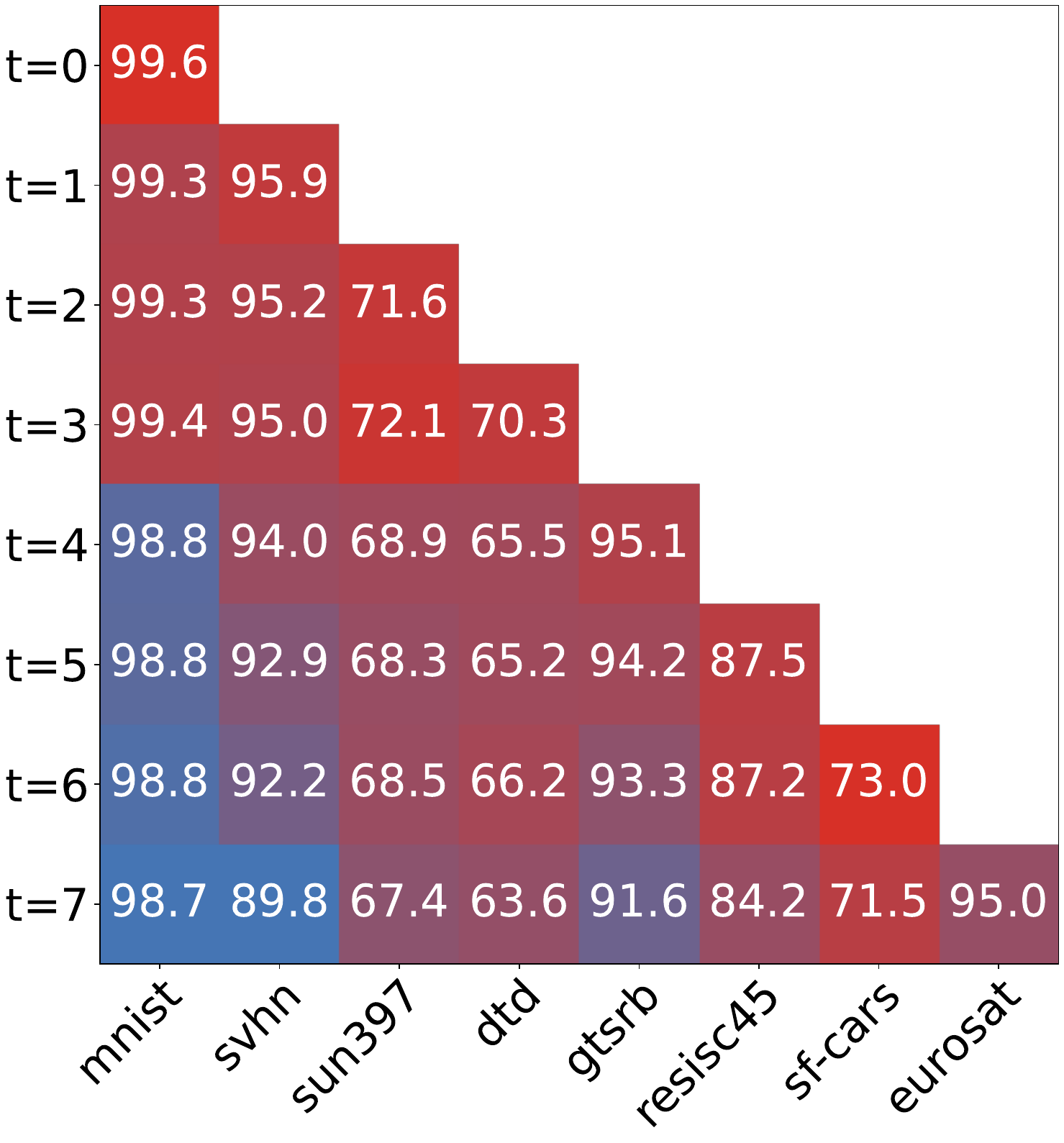}
        \caption{GS3R}
    \end{subfigure}
    \hspace{20pt}
    \begin{subfigure}{0.2\textwidth}
        \includegraphics[width=\linewidth]{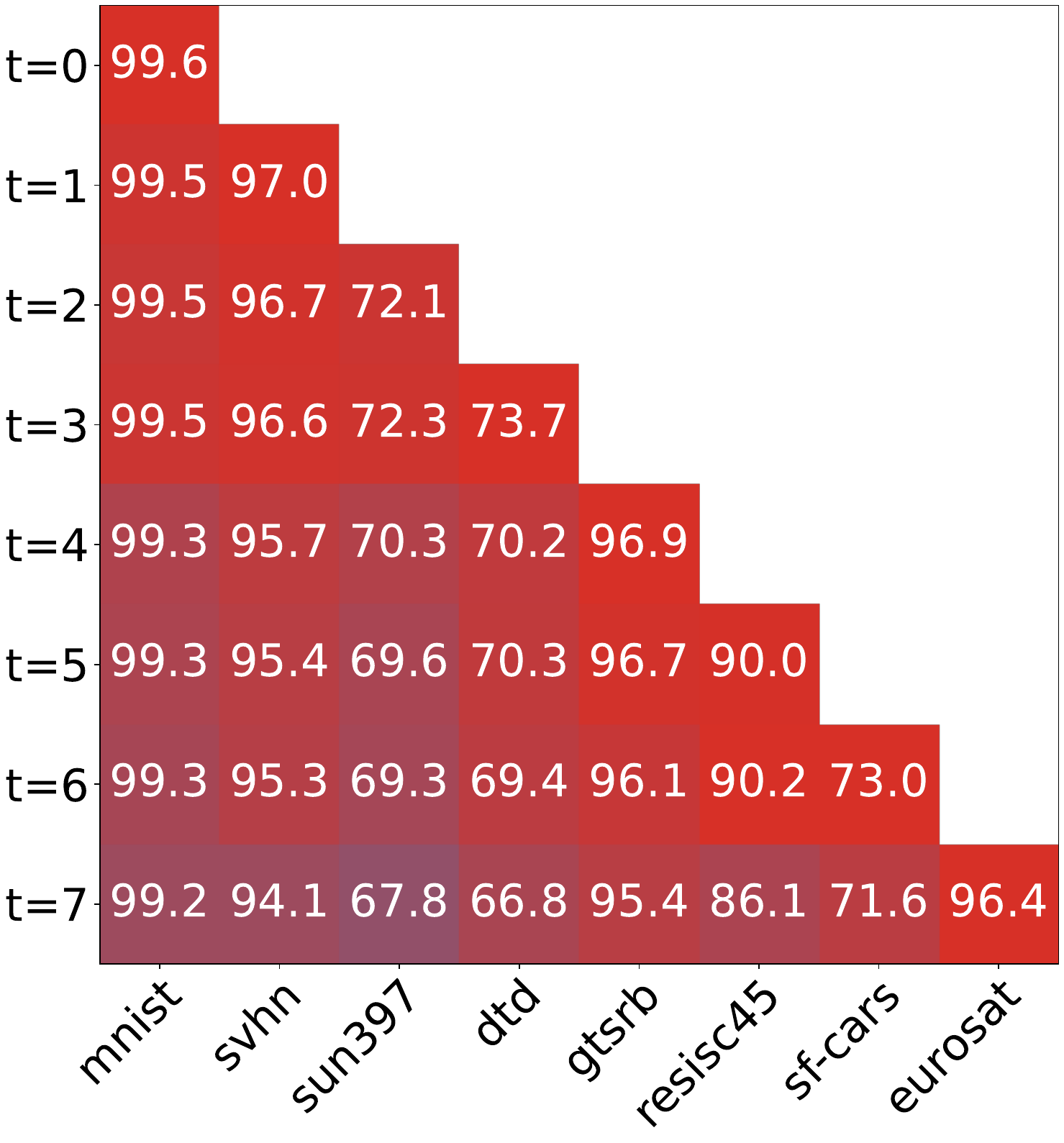}
        \caption{GS3R + Opt.}
    \end{subfigure}
    \caption{Accuracy trajectories during continual merging of ViT-B/32 over 8 vision tasks.
Rows correspond to merge steps and columns to tasks, showing how previously merged tasks evolve as new tasks arrive.
Colors are independently normalized per task.}

    \label{fig:heatmap}
    
\end{figure}

\subsection{Qualitative Analysis of Merging Behavior}

To visualize how task balance evolves during continual merging, we track the accuracy of each observed task after every merge step on the 8-task ViT-B/32 setting. Figure~\ref{fig:heatmap} reveals a clear difference in merging dynamics. DOP exhibits pronounced degradation on several earlier tasks as new tasks are incorporated, illustrating the task balance collapse discussed in Section~\ref{sec:collapse}. In contrast, NUFILT and GS3R both mitigate this degradation and better preserve the balance and accuracy of previously merged tasks.
This behavior is consistent with GS3R explicitly recovering and reallocating capacity across retained task-wise components at each merge step. 
With auxiliary optimization, GS3R further improves the final task accuracies while maintaining this balanced merging behavior.

Additional t-SNE~\citep{tsne} visualizations in Appendix~\ref{appendix:qual} provide a complementary representation-level view of the merged models. For a task merged early in a 20-task sequence, GS3R preserves well-separated and class-consistent clusters, whereas prior methods lose much of this structure.

\section{Conclusion}

We identified two coupled causes of task balance collapse in continual model merging: the entanglement of past task components and the accumulated distortion of their singular geometry. To address both, we proposed \emph{Global Singular Subspace Separation and Restoration (GS3R)}, which preserves a compact recoverable state alongside the merged update. This enables past task subspaces to be restored, separated, and globally reallocated at each merge without storing previous task models. Together with rank-controlled truncation and Pre-merge Restoration and Flushing (PRF), GS3R provides a practical accuracy--resource trade-off across vision, NLP, and multimodal benchmarks. More broadly, our results highlight that continual merging should preserve not only the current merged model, but also the structure required for future merges.

\begin{ack}
This work was supported by IITP and NRF grant funded by the Korea government (MSIT) (RS-2019-II191906, RS-2024-00415602, RS-2026-25490269).
\end{ack}

\bibliographystyle{unsrtnat}
\bibliography{main}


\newpage

\appendix

\section{Theoretical Proof of Restoration}
\label{appendix:proof}
In Section~\ref{sec:orthogonalization}, we explain the construction of restoration matrices. Here, we present the detailed restoration process and prove that the restoration matrices indeed recover the damage induced by orthogonalization (i.e., $\hat{U}^t = U_\perp^t R_U^t$ and $\hat{V}^t = V_\perp^t R_V^t$). We explain only the restoration of $\hat{U}^t$, as the exact same procedure is applied to $\hat{V}^t$.

\begin{proof}  
Let the SVD of $\hat{U}^t$ be:
\begin{align}
    \hat{U}^t
    =
    P_U^t D_U^t (Q_U^t)^\top,
    \qquad
    U_\perp^t
    =
    P_U^t (Q_U^t)^\top .
\end{align}
We define the restoration matrix as follows:
\begin{align}
    R_U^t
    =
    Q_U^t D_U^t (Q_U^t)^\top .
\end{align}
Equivalently, using only the orthogonalized basis and the SVD factors, this can be expressed as:
\begin{align}
    R_U^t
    =
    (U_\perp^t)^\top
    P_U^t D_U^t (P_U^t)^\top
    U_\perp^t .
\end{align}
Then the restoration guarantee follows directly as $(Q_U^t)^\top Q_U^t=I$:
\begin{align}
    U_\perp^t R_U^t 
    =
    P_U^t (Q_U^t)^\top
    Q_U^t D_U^t (Q_U^t)^\top
    =
    P_U^t D_U^t (Q_U^t)^\top
    =
    \hat{U}^t .
\end{align}
\end{proof}

\section{Experimental Details}
\label{appendix:exp}
We provide detailed experimental settings for merging vision, NLP, and multimodal tasks, including configurations for auxiliary optimization.

\subsection{Vision Task Orders}

The 8-task setting includes SUN397~\citep{sun}, Cars~\citep{cars}, RESISC45~\citep{resisc}, EuroSAT~\citep{eurosat}, SVHN~\citep{svhn}, GTSRB~\citep{gtsrb}, MNIST~\citep{mnist}, and DTD~\citep{dtd}. The 14-task setting further includes Flowers102~\citep{flowers}, PCAM~\citep{pcam}, FER2013~\citep{fer2013}, OxfordIIITPet~\citep{oxfordiiitpet}, STL10~\citep{stl10}, and CIFAR100~\citep{cifar}. The 20-task setting additionally includes CIFAR10~\citep{cifar}, Food101~\citep{food101}, FashionMNIST~\citep{fashionmnist}, EMNIST~\citep{emnist}, KMNIST~\citep{kmnist}, and RenderedSST2~\citep{clip, render1}. Task orders are generated using 10 seeds (40–49):
\paragraph{8-task orders}
\begin{itemize}
    \item MNIST, SVHN, SUN397, DTD, GTSRB, RESISC45, Cars, EuroSAT
    \item Cars, GTSRB, MNIST, SUN397, EuroSAT, RESISC45, DTD, SVHN
    \item EuroSAT, DTD, RESISC45, SUN397, SVHN, MNIST, GTSRB, Cars
    \item GTSRB, DTD, RESISC45, SUN397, MNIST, SVHN, EuroSAT, Cars
    \item SVHN, DTD, MNIST, GTSRB, Cars, SUN397, RESISC45, EuroSAT
    \item MNIST, DTD, SVHN, RESISC45, GTSRB, SUN397, EuroSAT, Cars
    \item MNIST, RESISC45, DTD, SUN397, Cars, SVHN, GTSRB, EuroSAT
    \item Cars, GTSRB, RESISC45, DTD, SUN397, SVHN, EuroSAT, MNIST
    \item MNIST, SUN397, RESISC45, SVHN, Cars, DTD, GTSRB, EuroSAT
    \item SVHN, MNIST, EuroSAT, Cars, RESISC45, DTD, GTSRB, SUN397
    
\end{itemize}

\paragraph{14-task orders}
\begin{itemize}
    \item OxfordIIITPet, STL10, PCAM, Cars, Flowers102, SUN397, MNIST, DTD, SVHN, GTSRB, RESISC45, EuroSAT, CIFAR100, FER2013
    \item RESISC45, FER2013, PCAM, Cars, DTD, CIFAR100, GTSRB, EuroSAT, SUN397, MNIST, STL10, Flowers102, SVHN, OxfordIIITPet
    \item EuroSAT, GTSRB, RESISC45, SVHN, PCAM, CIFAR100, DTD, SUN397, MNIST, STL10, Flowers102, FER2013, Cars, OxfordIIITPet
    \item Flowers102, CIFAR100, GTSRB, RESISC45, MNIST, EuroSAT, SVHN, SUN397, FER2013, PCAM, DTD, Cars, STL10, OxfordIIITPet
    \item Flowers102, MNIST, GTSRB, FER2013, RESISC45, STL10, DTD, PCAM, OxfordIIITPet, Cars, CIFAR100, SUN397, SVHN, EuroSAT
    \item RESISC45, Flowers102, EuroSAT, STL10, OxfordIIITPet, GTSRB, MNIST, PCAM, FER2013, SVHN, CIFAR100, SUN397, DTD, Cars
    \item Cars, SVHN, CIFAR100, GTSRB, STL10, MNIST, DTD, SUN397, OxfordIIITPet, PCAM, RESISC45, Flowers102, FER2013, EuroSAT
    \item Cars, GTSRB, SVHN, EuroSAT, RESISC45, CIFAR100, DTD, FER2013, OxfordIIITPet, SUN397, PCAM, STL10, MNIST, Flowers102
    \item FER2013, SUN397, CIFAR100, SVHN, DTD, GTSRB, STL10, Cars, PCAM, RESISC45, MNIST, OxfordIIITPet, EuroSAT, Flowers102
    \item CIFAR100, MNIST, DTD, Cars, PCAM, GTSRB, EuroSAT, RESISC45, SVHN, OxfordIIITPet, FER2013, SUN397, STL10, Flowers102
\end{itemize}

\paragraph{20-task orders}
\begin{itemize}
    \item STL10, KMNIST, GTSRB, FER2013, Cars, PCAM, RenderedSST2, DTD, EuroSAT, Food101, RESISC45, EMNIST, FashionMNIST, SUN397, CIFAR100, CIFAR10, Flowers102, OxfordIIITPet, SVHN, MNIST
    \item MNIST, CIFAR10, CIFAR100, OxfordIIITPet, SVHN, KMNIST, Cars, GTSRB, RenderedSST2, RESISC45, DTD, Food101, FER2013, EuroSAT, FashionMNIST, SUN397, STL10, Flowers102, EMNIST, PCAM
    \item KMNIST, FashionMNIST, FER2013, DTD, SVHN, MNIST, GTSRB, OxfordIIITPet, CIFAR10, PCAM, Flowers102, STL10, RenderedSST2, Food101, SUN397, Cars, CIFAR100, RESISC45, EMNIST, EuroSAT
    \item Cars, OxfordIIITPet, SVHN, RenderedSST2, EMNIST, DTD, STL10, FashionMNIST, RESISC45, FER2013, MNIST, Flowers102, KMNIST, GTSRB, PCAM, CIFAR100, SUN397, Food101, CIFAR10, EuroSAT
    \item Food101, STL10, CIFAR100, CIFAR10, OxfordIIITPet, GTSRB, FashionMNIST, FER2013, MNIST, SVHN, Flowers102, RESISC45, Cars, EMNIST, KMNIST, EuroSAT, RenderedSST2, SUN397, PCAM, DTD
    \item KMNIST, FashionMNIST, MNIST, CIFAR10, Flowers102, GTSRB, DTD, RenderedSST2, CIFAR100, EMNIST, FER2013, OxfordIIITPet, SVHN, Cars, STL10, PCAM, EuroSAT, SUN397, Food101, RESISC45
    \item MNIST, CIFAR10, CIFAR100, EuroSAT, PCAM, Flowers102, RenderedSST2, EMNIST, RESISC45, OxfordIIITPet, Food101, FER2013, GTSRB, STL10, KMNIST, SUN397, Cars, SVHN, FashionMNIST, DTD
    \item EMNIST, GTSRB, PCAM, EuroSAT, CIFAR10, RESISC45, MNIST, OxfordIIITPet, Flowers102, DTD, SVHN, Food101, RenderedSST2, SUN397, FER2013, CIFAR100, Cars, KMNIST, STL10, FashionMNIST
    \item CIFAR10, KMNIST, RenderedSST2, Food101, SUN397, OxfordIIITPet, Flowers102, RESISC45, PCAM, EuroSAT, DTD, Cars, FER2013, EMNIST, STL10, SVHN, GTSRB, CIFAR100, MNIST, FashionMNIST
    \item EMNIST, GTSRB, MNIST, STL10, PCAM, RESISC45, CIFAR10, Food101, EuroSAT, CIFAR100, RenderedSST2, FER2013, SUN397, OxfordIIITPet, DTD, SVHN, Flowers102, KMNIST, Cars, FashionMNIST
\end{itemize}

\subsection{NLP Task Orders}
We conduct NLP experiments on eight tasks from the GLUE benchmark~\citep{glue}, namely CoLA, MNLI, MRPC, QNLI, QQP, RTE, SST2, and STSB. For STSB, we report Spearman’s correlation $\rho$. Task orders are generated using 10 seeds (40–49):
\paragraph{GLUE}
\begin{itemize}
    \item MRPC, RTE, QNLI, MNLI, CoLA, SST2, QQP, STSB
    \item CoLA, QQP, STSB, QNLI, RTE, MNLI, MRPC, SST2
    \item QNLI, QQP, SST2, STSB, MRPC, RTE, CoLA, MNLI
    \item SST2, STSB, QQP, QNLI, MNLI, RTE, MRPC, CoLA
    \item MRPC, RTE, QNLI, MNLI, CoLA, STSB, QQP, SST2
    \item RTE, STSB, MNLI, CoLA, MRPC, SST2, QNLI, QQP
    \item SST2, RTE, MRPC, STSB, QQP, CoLA, QNLI, MNLI
    \item SST2, MNLI, MRPC, STSB, QQP, QNLI, CoLA, RTE
    \item SST2, CoLA, QNLI, MRPC, STSB, QQP, MNLI, RTE
    \item STSB, QQP, RTE, SST2, CoLA, QNLI, MRPC, MNLI
\end{itemize}

\subsection{Multimodal Task Orders}

We conduct multimodal experiments on four vision-language tasks, namely IconQA~\citep{iconqa}, ImageNet~\citep{imagenet}, ScienceQA~\citep{scienceqa}, and VizWiz~\citep{vizwiz}, using the checkpoints and following the evaluation protocol of RobustMerge~\citep{RobustMerge}. In addition to the alphabetical order, task orders are generated using 5 seeds (40--43, 45). Seed 44 is skipped since it yields the same permutation as seed 40, so that all six orders are distinct:
\begin{itemize}
    \item IconQA, ImageNet, ScienceQA, VizWiz
    \item ImageNet, IconQA, ScienceQA, VizWiz
    \item ScienceQA, IconQA, ImageNet, VizWiz
    \item ScienceQA, ImageNet, VizWiz, IconQA
    \item ScienceQA, VizWiz, ImageNet, IconQA
    \item IconQA, VizWiz, ImageNet, ScienceQA
\end{itemize}

\subsection{Hyperparameters}

\paragraph{Global singular subspace separation} We use a fixed orthogonality tolerance of $\epsilon = 10^{-4}$ throughout all experiments.

\paragraph{Auxiliary optimization} For each layer, we optimize the merged weights using the Adam optimizer with a learning rate of $1 \times 10^{-5}$ for vision and multimodal tasks and $1 \times 10^{-4}$ for NLP tasks. For ViT variants and LLaVA, all matrix-shaped weights are optimized, whereas for the Flan-T5 architecture, embedding layers are excluded. We observe that the number of required optimization steps increases with the number of tasks. Accordingly, we perform 300, 400, 800, and 1000 iterations when merging 4, 8, 14, and 20 tasks, respectively.

\paragraph{Multimodal baselines} For multimodal tasks, Continual Task Arithmetic and Continual TIES-Merging use a scaling coefficient of $\lambda=0.3$, with TIES-Merging keeping the top 20\% of parameters, and OPCM uses its default setting of $\alpha=0.5$ in FusionBench~\citep{FusionBench}. NUFILT is run with its official implementation and default hyperparameters ($r_p=128$, $r_l=64$, $r_v=8$, 50 iterations, learning rate $1\times10^{-3}$). Only the task order is varied, identically for all methods.

\section{Additional Analysis and Experimental Results}

\subsection{Inter- and Intra-task Orthogonality of Singular Vectors}
\label{appendix:orth}

\begin{figure}[H]
     \centering
         \includegraphics[width=1.0\columnwidth]{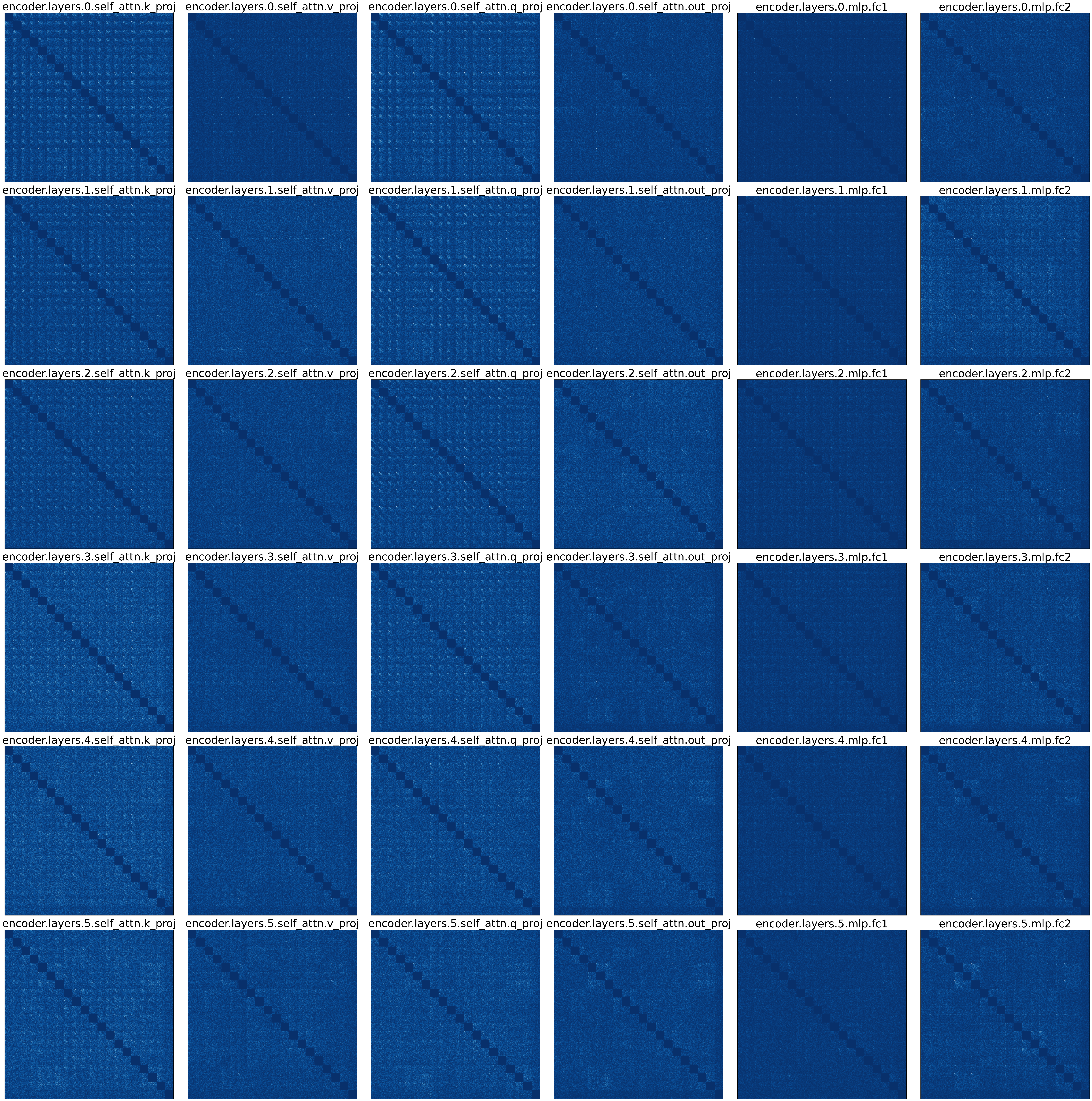}
     \caption{Gram matrix of concatenated singular vectors from $U$ of each task.}

     \label{fig:Gram_U}
\end{figure}

\begin{figure}[H]
     \centering
         \includegraphics[width=1.0\columnwidth]{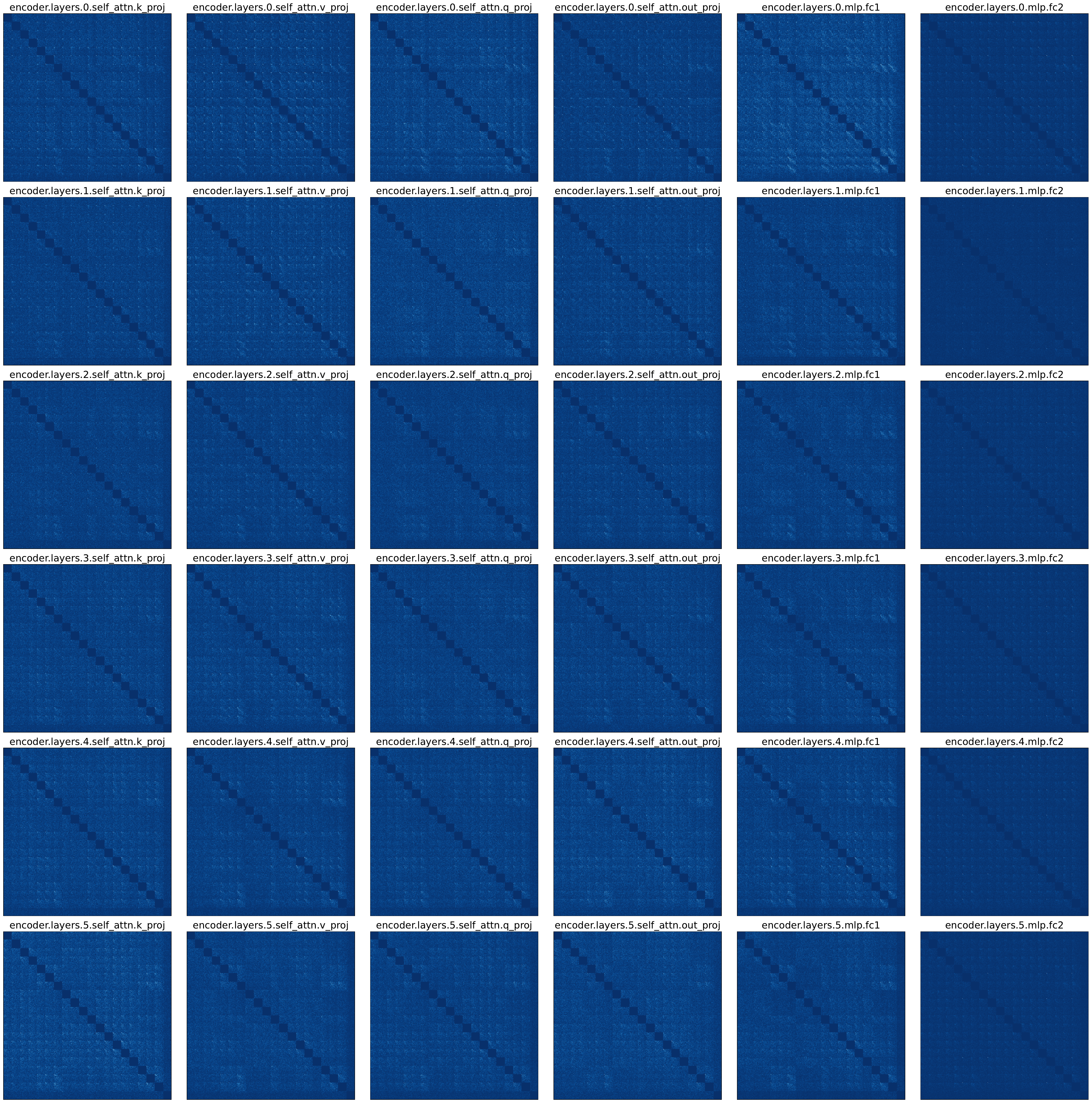}
     \caption{Gram matrix of concatenated singular vectors from $V$ of each task.}

     \label{fig:Gram_V}
\end{figure}

The subspace separation in GS3R is based on the observation that singular vectors within a single task are mutually orthogonal, whereas orthogonality across different tasks does not necessarily hold. To demonstrate this, for each weight matrix, we collect the corresponding singular vectors from every task and concatenate them according to the shape of the weight. We then compute the Gram matrices of these concatenated singular vectors and present them in Figure~\ref{fig:Gram_U} and Figure~\ref{fig:Gram_V}.

The Gram matrices from different layers exhibit a clear block-diagonal structure, where each diagonal block corresponds to the singular vectors of an individual task. This pattern confirms that intra-task orthogonality is preserved, while inter-task orthogonality is not. Such a structural property justifies our subspace separation strategy, as it allows GS3R to effectively isolate task-specific subspaces via spectral clustering.

\vspace*{\fill}
\vspace*{\fill}

\clearpage
\vspace*{\fill}
\vspace*{\fill}

\clearpage

\subsection{Details and Analysis on Spectral Clustering}
\label{appendix:spectral_clustering}

In this section, we further describe and analyze the spectral clustering used in the subspace separation step. Specifically, we provide its algorithmic details, validate its reliability, and present an ablation study on the tolerance parameter $\epsilon$.

\paragraph{Algorithmic details}

Algorithm~\ref{alg:separation} summarizes the separation step. The key property it exploits is that singular vectors originating from the same task are mutually orthogonal by construction of the SVD, while vectors from different tasks typically carry a small but nonzero correlation. The affinity therefore encodes pairwise orthogonality. Two normalized columns are connected ($A_{ij} = 1$) when their absolute inner product falls below the tolerance $\epsilon$, so each task's components form a densely connected group (lines 4--12). We then form the symmetric normalized Laplacian and compute its eigenvalues in ascending order (lines 13--15). When the graph consists of $k$ well-separated dense groups, the smallest $k$ eigenvalues concentrate near zero and a clear jump appears afterward, so the number of tasks is selected as the position of the largest gap between consecutive eigenvalues (line 16). Note that the number of merged tasks is never supplied as an input; it is inferred entirely from the accumulated structure. Clustering is completed by row-normalizing the first $k$ eigenvectors and running $k$-means, following standard normalized spectral clustering, and the components of each recovered group are sorted by singular value for the subsequent rank selection (lines 17--25).

\begin{algorithm}[t]
\caption{Subspace Separation via Spectral Clustering}
\label{alg:separation}
\begin{algorithmic}[1]
\Require Restored singular vectors $U = [u_1, \dots, u_N]$, singular values $\sigma_1, \dots, \sigma_N$, tolerance $\epsilon$
\For{$i = 1, \dots, N$} \Comment{Normalize Columns}
    \State $\bar{u}_i \gets u_i / \lVert u_i \rVert$
\EndFor
\For{$i = 1, \dots, N$} \Comment{Build Orthogonality Affinity}
    \For{$j = 1, \dots, N$}
        \If{$i \neq j$ \textbf{and} $|\langle \bar{u}_i, \bar{u}_j \rangle| < \epsilon$}
            \State $A_{ij} \gets 1$
        \Else
            \State $A_{ij} \gets 0$
        \EndIf
    \EndFor
\EndFor
\State $D \gets \mathrm{diag}(A \cdot \mathbf{1})$ \Comment{Degree Matrix, $\mathbf{1}$ is the all-ones vector}
\State $L \gets I - D^{-1/2} A D^{-1/2}$ \Comment{Symmetric Normalized Laplacian}
\State $(\lambda_1 \leq \dots \leq \lambda_N),\, Q \gets \mathrm{eigendecomposition}(L)$
\State $k \gets \arg\max_{i} \left( \lambda_{i+1} - \lambda_i \right)$ \Comment{Eigengap Heuristic}
\State $Y \gets$ first $k$ columns of $Q$
\For{$i = 1, \dots, N$} \Comment{Row-normalize Embedding}
    \State $Y_{i,:} \gets Y_{i,:} / \lVert Y_{i,:} \rVert$
\EndFor
\State $\text{labels} \gets \mathrm{k\text{-}means}(Y, k)$
\For{$c = 1, \dots, k$} \Comment{Form Task-wise Groups}
    \State $G_c \gets \{\, i \mid \text{labels}_i = c \,\}$
    \State Sort $G_c$ by $\sigma_i$ in descending order
\EndFor
\State \Return $G_1, \dots, G_k$ and $k$
\end{algorithmic}
\end{algorithm}

\paragraph{Validation of clustering}

The separation criterion could in principle fail in two ways: components of different tasks may happen to be orthogonal within the tolerance, or related tasks may share dominant singular directions. To measure how often either occurs, we run the separation step on every weight matrix of ViT-B/32 (73 layers) with all 20 task vectors, and compare the recovered grouping against the ground-truth task identity via the correct cluster count ratio and the Adjusted Rand Index (ARI). The 20 task suite is the relevant stress case for the second failure mode, as it contains four MNIST-family datasets, SVHN, and both CIFAR-10 and CIFAR-100. As shown in Table~\ref{tab:clustering_ari}, at the default tolerance ($\epsilon = 10^{-4}$) separation is exact in every layer. Notably, shared dominant directions produce no misassignment despite the closely related datasets, consistent with the block-diagonal Gram matrices in Appendix~\ref{appendix:orth}. Degradation appears only at $\epsilon = 10^{-2}$, where the tolerance approaches the typical magnitude of cross-task inner products and different tasks begin to be linked in the affinity graph.

\begin{table}[t]
\centering
\caption{Reliability of the separation step on ViT-B/32 with 20 tasks under varying tolerance $\epsilon$. We report the fraction of layers (out of 73) with the correct cluster count and the mean Adjusted Rand Index. $\epsilon = 10^{-4}$ is used throughout the paper.}
\label{tab:clustering_ari}
\begin{tabular}{c|ccccc}
\toprule
Tolerance $\epsilon$ & $10^{-2}$ & $10^{-3}$ & $10^{-4}$ & $10^{-5}$ & $10^{-6}$ \\
\midrule
Correct cluster count & 0/73 & 70/73 & \textbf{73/73} & \textbf{73/73} & \textbf{73/73} \\
Mean ARI & 0.007 & 0.959 & \textbf{1.000} & \textbf{1.000} & \textbf{1.000} \\
\bottomrule
\end{tabular}
\end{table}

\paragraph{Ablation study on tolerance parameter $\epsilon$}

\begin{table}[t]
\centering
\caption{Effect of the tolerance $\epsilon$ on merging performance, merging 8 tasks with ViT-B/32 over 10 task orders.}
\label{tab:eps_sweep}
\begin{tabular}{c|ccccc}
\toprule
Tolerance $\epsilon$ & $10^{-2}$ & $5\times10^{-3}$ & $10^{-3}$ & $10^{-4}$ & $10^{-5}$ \\
\midrule
Accuracy (\%) & 81.6$\pm$0.6 & 82.4$\pm$0.3 & \textbf{82.7}$\pm$0.2 & \textbf{82.7}$\pm$0.2 & \textbf{82.7}$\pm$0.2 \\
\bottomrule
\end{tabular}
\end{table}

Table~\ref{tab:eps_sweep} reports the merging performance when sweeping $\epsilon$ over three orders of magnitude (8 tasks, ViT-B/32, 10 task orders). Accuracy is flat across the range where the separation is exact and drops only as $\epsilon$ approaches the cross-task inner-product scale, the same point where the clustering itself breaks down. The default value thus sits in the middle of a wide plateau rather than at a tuned point.

\subsection{Analysis of Full-rank Degradation of NLP Tasks}

\label{appendix:rank_analysis}

As reported in Section~\ref{sec:nlp_exp}, the truncated variant of GS3R outperforms its full-rank counterpart on GLUE, while the ordering is reversed on the vision suite. We first verify that this is not a seed-level fluctuation. The rank sweep with 10 new task orders (Table~\ref{tab:rank_sweep_seeds}) reproduces the same pattern, with accuracy peaking at rank budget $r=64$ and falling off toward both ends by margins an order of magnitude larger than the seed variance. The trend is also not monotonic in $r$, so a single error source cannot account for it. We identify and measure two sources separately below.

\begin{table}
\centering
\caption{Rank sweep on GLUE with Flan-T5-Base, re-run with 10 new task orders. The same pattern as the original experiment appears: accuracy peaks at $r=64$ and falls off toward both ends by margins an order of magnitude larger than the seed variance.}
\label{tab:rank_sweep_seeds}
\scalebox{0.8}{
    \begin{tabular}{c|cccccc}
    \toprule
    $r$ & 16 & 32 & 64 & 128 & 256 & Full Rank \\
    \midrule
    Accuracy (\%) & 82.73$\pm$0.11 & 83.59$\pm$0.10 & \textbf{83.63}$\pm$0.06 & 83.38$\pm$0.12 & 83.30$\pm$0.08 & 82.64$\pm$0.17 \\
    \bottomrule
    \end{tabular}
    }
\end{table}

\paragraph{Effective rank differs sharply by architecture}

We measure the energy rank of the task vectors, defined as the minimum rank capturing 90\% of $\sum_i \sigma_i^2$, averaged over tasks and layers (Table~\ref{tab:energy_rank}). Encoder updates in Flan-T5 are no more concentrated than the vision suite (197.6 vs.\ 147.2), but decoder updates saturate at a far smaller rank (26.1). Since decoder matrices form the majority of the merged matrices in Flan-T5, their early saturation pulls the aggregate down (93.0) and governs the behavior under a uniform budget.

\begin{table}[t]
\centering
\caption{Energy rank of task vectors, defined as the minimum rank capturing 90\% of $\sum_i \sigma_i^2$, averaged over tasks and layers.}
\label{tab:energy_rank}
\begin{tabular}{c|cccc}
\toprule
& \multicolumn{3}{c}{GLUE: Flan-T5} & Vision suite \\
\cmidrule(lr){2-4} \cmidrule(lr){5-5}
& Decoder & Encoder & All & ViT-B/32 (8 tasks) \\
\midrule
Energy rank & 26.1 & 197.6 & 93.0 & 147.2 \\
\bottomrule
\end{tabular}
\end{table}

\paragraph{Error decomposition}

To locate where the damage occurs, we decompose the merging error of each task vector $\tau = U \Sigma V^\top$ into two stages, measured task-wise per layer. Truncation first reduces $\tau$ to the $r/N$ directions allocated to each of the $N$ tasks under the total budget $r$, written $\tau' = U' \Sigma' (V')^\top$. Orthogonalization then stacks the truncated bases across tasks and replaces them with their polar factors, holding the singular values fixed. We write $U_\perp$ and $V_\perp$ for the resulting blocks of this task. The two stages are measured as follows:

\begin{equation}
\label{eq:error_decomp}
E_\mathrm{trunc} = 1 - \frac{\lVert \tau' \rVert_F}{\lVert \tau \rVert_F}, \qquad
E_\mathrm{orth} = \frac{\lVert U_\perp \Sigma' V_\perp^\top - \tau' \rVert_F}{\lVert \tau' \rVert_F}, \qquad
E_\mathrm{total} = \frac{\lVert U_\perp \Sigma' V_\perp^\top - \tau \rVert_F}{\lVert \tau \rVert_F}.
\end{equation}

Because $\Sigma'$ appears identically in both terms of $E_\mathrm{orth}$, it isolates the displacement of the singular bases caused by orthogonalization alone. The three quantities are measured independently and are not additive, since their reference points differ.

\paragraph{Orthogonalization dominates on GLUE, truncation on vision}

Table~\ref{tab:error_decomp} shows that the two suites sit in opposite regimes. On GLUE, $E_\mathrm{trunc}$ is already low at the smallest budget and falls only slowly, consistent with the low effective rank of the decoder, while $E_\mathrm{orth}$ is the larger term throughout and grows with $r$. The total is therefore minimized at an intermediate budget, matching the optimum in Figure~\ref{fig:trade-off}(a). On the vision suite the ordering reverses. Truncation dominates and drops steeply as $r$ grows, since the task vectors still carry signal in those directions, so the total decreases monotonically and full rank performs best. The same measurement predicting opposite optima on the two suites, both agreeing with the measured accuracy, supports its validity as a diagnostic.

\begin{table}[t]
\centering
\caption{Error decomposition under varying total rank budget $r$. $E_\mathrm{trunc}$, $E_\mathrm{orth}$, and $E_\mathrm{total}$ are defined in Eq. (\ref{eq:error_decomp}). On GLUE, orthogonalization damage dominates and grows with $r$, so the total is minimized at an intermediate budget. On the vision suite, truncation dominates and the total decreases monotonically.}
\label{tab:error_decomp}
\begin{tabular}{cc|ccccc}
\toprule
& $r$ & 16 & 32 & 64 & 128 & 256 \\
\midrule
\multirow{3}{*}{GLUE}
& $E_\mathrm{trunc}$ & 0.30 & 0.23 & 0.18 & 0.14 & 0.10 \\
& $E_\mathrm{orth}$  & 0.67 & 0.72 & 0.76 & 0.81 & 0.87 \\
& $E_\mathrm{total}$ & 0.87 & \textbf{0.86} & \textbf{0.86} & 0.88 & 0.92 \\
\midrule
\multirow{3}{*}{Vision}
& $E_\mathrm{trunc}$ & 0.59 & 0.50 & 0.40 & 0.30 & 0.21 \\
& $E_\mathrm{orth}$  &0.47 & 0.48 & 0.52 & 0.58 & 0.67 \\
& $E_\mathrm{total}$ & 0.93 & 0.90 & 0.86 & \textbf{0.82} & \textbf{0.82} \\
\bottomrule
\end{tabular}
\end{table}

\paragraph{Why surplus directions are harmful}

Directions beyond the signal rank are not inert. During orthogonalization they occupy the shared basis space and displace the signal directions of other tasks, which is precisely why $E_\mathrm{orth}$ grows with $r$. Truncation acts as a filter on these surplus directions, whereas orthogonalization converts them into signal displacement, which is why full rank underperforms when the effective rank is small.

\subsection{Evaluation Protocol for Multimodal Tasks}
\label{appendix:llava_eval}

\paragraph{Official scoring rules}
The four benchmarks are scored by the evaluation scripts released with RobustMerge~\citep{RobustMerge}, and three of them are tolerant to the formatting of generated text by construction. ImageNet counts a prediction as correct when the ground-truth label and the prediction contain each other as substrings, ScienceQA extracts the selected option from the generated sentence with a regular expression before comparison, and VizWiz-Caption averages standard captioning metrics. IconQA is the only exception. Its scorer counts a prediction as correct only when the full generated string exactly matches the ground-truth answer string after case normalization, so any additional character makes an otherwise correct answer count as wrong.

\paragraph{Observed failure mode}
Inspecting the IconQA outputs of the merged models, we found that the exact-match scorer primarily penalizes formatting drift rather than wrong answers. Merged models frequently append a trailing period or additional tokens to the answer, such as ``4.'' for the ground truth ``4'' or ``0. paw.'' for ``0'', where the selected option is correct but the string comparison fails. The individually fine-tuned checkpoint produces no such outputs, so the drift is introduced by merging rather than inherited from fine-tuning. It appears in every merging method, only marginally for Continual Task Arithmetic. Ignoring the trailing text raises IconQA accuracy by 6 to 13 points for Continual TIES-Merging, OPCM, and the GS3R variants, and by 24 points for NUFILT, whose exact-match score also varies strongly across task orders.

\paragraph{Format-tolerant rule}
To evaluate IconQA at the same level of tolerance as the other three tasks, we additionally count a prediction as correct if its first token (the leading characters up to whitespace, a period, or a comma), or the full string after stripping a trailing period, matches the ground-truth answer, case-insensitively. The rule is applied identically to all methods and changes only the IconQA column. Table~\ref{tab:llava} reports both the original (ACC) and the format-tolerant (ACC$^\dagger$) averages, with the format-tolerant IconQA accuracy given in parentheses.

\paragraph{Discussion}
The two scoring rules measure different things and lead to the same conclusion. The original scorer folds formatting drift into the IconQA score, and under it GS3R outperforms all baselines, with part of this margin reflecting that the merged model of NUFILT drifts further from the expected answer format than ours. The format-tolerant rule isolates whether the correct answer was produced, and under it truncated GS3R matches NUFILT without any optimization while GS3R with optimization remains the best method. GS3R therefore matches or surpasses NUFILT under both protocols, and unlike NUFILT its results remain stable across task orders under exact match. We report both so that the contribution of formatting is visible rather than hidden in the average. How different merging operators affect the output format of generative models is an interesting question in its own right, which we leave to future work.

\subsection{Rank Allocation Strategy}

In our implementation, GS3R treats all tasks equally and allocates the rank budget uniformly across tasks. However, since all task-specific information is separated in the separation step, different rank allocation strategies can be employed. Beyond the uniform allocation used throughout the paper, we evaluate four alternatives that allocate the rank budget from task-specific properties of the spectrum. Three allocate proportionally to nuclear mass, energy, and Frobenius norm, respectively, while the fourth is a global top-$k$ selection that ignores task boundaries entirely. Table~\ref{tab:rank_alloc} reports the results of merging 8 tasks with ViT-B/32 over 10 task orders.

\begin{table}[h]
\centering
\caption{Comparison of rank allocation strategies when merging 8 tasks with ViT-B/32. We report mean and standard deviation of accuracy over 10 task orders.}
\label{tab:rank_alloc}
\begin{tabular}{c|ccccc}
\toprule
& Nuclear & Energy & Frobenius & Global Top-$k$ & Ours \\
\midrule
Accuracy (\%) & 82.3$\pm$0.3 & 82.3$\pm$0.2 & 82.5$\pm$0.3 & \underline{82.6}$\pm$0.2 & \textbf{82.7}$\pm$0.2 \\
\bottomrule
\end{tabular}
\end{table}

None of the alternatives improves on uniform allocation. Since each variant reads intrinsic rank differently from the same spectrum, this indicates that the information needed for a better allocation is not present in the task vectors themselves. Task complexity and importance are properties of the deployment scenario rather than of the weight space, so a policy exploiting them would require a signal from outside the data-free setting. Because the allocation rule only determines which indices are selected in Eq. (\ref{eq:selected_V}), such a policy can replace ours without modifying the rest of the pipeline, and we leave this as a direction for future work.

\subsection{Comparison with Explicit Subspace Storage}

One simple yet intuitive baseline not considered in the main paper is to explicitly store the task-wise subspaces. For a layer with task matrix $\tau \in \mathbb{R}^{m\times n}$ and rank budget $r$, this requires storing $(m+n)r$ entries per task, while GS3R stores only $r^2$ entries regardless of the model width. The advantage grows as the rank budget decreases, which is precisely the regime of edge systems.

The results are summarized in Table~\ref{tab:explicit_storage}. Even at full rank, GS3R uses about $3\times$ less storage, and the gap widens to $9$--$69\times$ in the truncated regime, since the metadata size of GS3R depends only on the rank budget and not on the model width. At the same time, the accuracy difference is under 0.5\%p at every rank. Since the baseline stores the exact per-task subspaces, it serves as an upper bound where task identity and pre-orthogonalization geometry are given for free. That GS3R recovers this bound within 0.5\%p, without storing any task identity, is direct evidence that restoration and separation are practically effective.

\begin{table}
\centering
\caption{Comparison with the explicit-storage baseline under varying rank budgets, merging 8 tasks with ViT-B/32. We report the mean accuracy (\%) and the additional storage (MB). Results of GS3R are averaged over 10 task orders. The explicit-storage baseline is deterministic given the task order.}
\label{tab:explicit_storage}
\begin{tabular}{ccccc}
\toprule
& \multicolumn{2}{c}{Accuracy (\%)} & \multicolumn{2}{c}{Storage (MB)} \\
\cmidrule(lr){2-3} \cmidrule(lr){4-5}
Rank Budget ($r$) & Explicit & GS3R & Explicit & GS3R \\
\midrule
32   & 72.9 & 72.7$\pm$0.3 & 21.3  & \textbf{0.3} \\
64   & 76.3 & 76.0$\pm$0.3 & 42.6  & \textbf{1.2} \\
128  & 79.4 & 79.1$\pm$0.2 & 85.1  & \textbf{4.8} \\
256  & 82.0 & 81.8$\pm$0.2 & 170.0 & \textbf{19.0} \\
Full & 83.1 & 82.7$\pm$0.2 & 509.8 & \textbf{170.1} \\
\bottomrule
\end{tabular}
\end{table}

\subsection{The Case of Repeated Tasks}
\label{appendix:repeated_tasks}

An interesting corner case of continual merging is when the same task arrives more than once. Since spectral clustering is applied to the previously merged vector and the incoming task is added afterward as a separate group, a task arriving a second time enters as an additional group and receives twice the rank budget. What is less obvious is what happens during orthogonalization once its components appear twice. We provide a theoretical analysis and empirical verification below.

\paragraph{Theoretical analysis}

Let $\hat{U}$ denote the concatenated left singular vectors of all selected components across tasks where each column is a unit vector. The analysis for $\hat{V}$ is identical. Orthogonalization replaces $\hat{U} = PDQ^\top$ with its polar factor $PQ^\top$. This resets every stretch value in $D$ to one. For a unit direction $w$, the stretch $\sigma_w^2 = \sum_c \langle w, u_c \rangle^2$ is the total load placed on $w$ by all columns of all tasks, so directions shared by several tasks are suppressed more strongly. On a direction $w$ shared by several tasks we split this sum into the terms coming from the columns of task $j$ and the terms coming from all other columns. Writing $\sigma_j^2 = \sum_{c \in j} \langle w, u_c \rangle^2$ and $\sigma_\mathrm{rest}^2 = \sum_{c \notin j} \langle w, u_c \rangle^2$ gives $\sigma_w^2 = \sigma_j^2 + \sigma_\mathrm{rest}^2$. The fraction $\alpha_j$ of that direction retained by task $j$ is

\begin{equation}
\alpha_j = \frac{\sigma_j^2}{\sigma_j^2 + \sigma_\mathrm{rest}^2} < 1.
\end{equation}

If the same task arrives $m$ times, its contribution becomes $m\sigma_j^2$ while $\sigma_\mathrm{rest}^2$ stays the same, and the retained fraction becomes

\begin{equation}
\alpha_j(m) = \frac{m\sigma_j^2}{m\sigma_j^2 + \sigma_\mathrm{rest}^2}.
\end{equation}

This increases with $m$ but never exceeds 1. Repetition therefore only restores what the overlap had discounted and never strengthens the task beyond what it would have on an unshared direction. The shares of the other tasks shrink accordingly, making this a reallocation of capacity rather than a corruption of the merge.

\paragraph{Empirical verification}

We verify this prediction in the full pipeline. After merging 8 tasks with ViT-B/32, the task appearing fourth in the order is merged once more. Order 1 is MNIST, SVHN, SUN397, DTD, GTSRB, RESISC45, Cars, EuroSAT, and Order 2 is MNIST, SUN397, RESISC45, SVHN, Cars, DTD, GTSRB, EuroSAT, so the repeated task is DTD and SVHN, respectively. As shown in Table~\ref{tab:repeated_tasks}, the repeated task improves substantially in both orders as predicted. As the results also show, however, this does not necessarily translate into an overall improvement since the shares of the remaining tasks shrink. Filtering component pairs with inner products near one removes this behavior if undesired, though a repeated arrival may itself signal that the task matters more in deployment.

\begin{figure}
    \centering
    \begin{subfigure}{0.23\textwidth}
        \includegraphics[width=\linewidth]{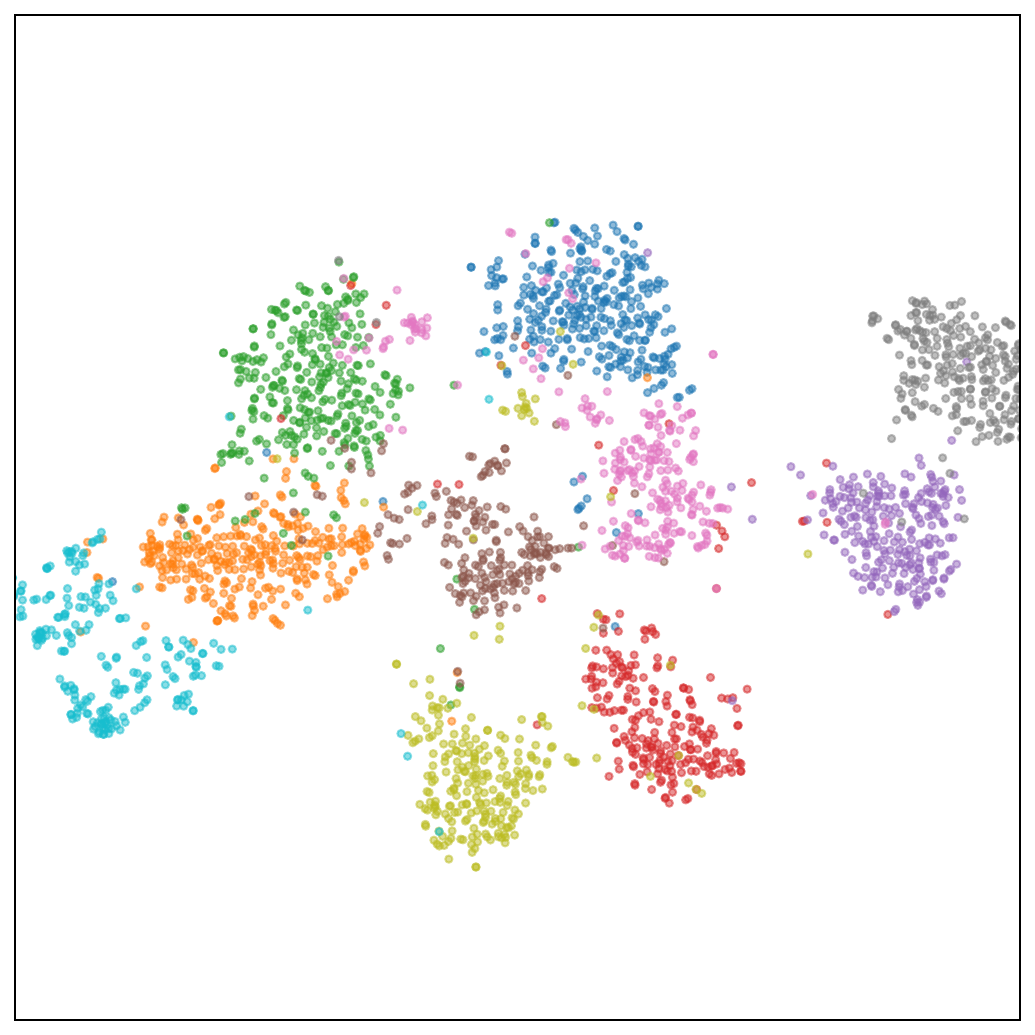}
        \caption{OPCM}
    \end{subfigure}\hfill
    \begin{subfigure}{0.23\textwidth}
        \includegraphics[width=\linewidth]{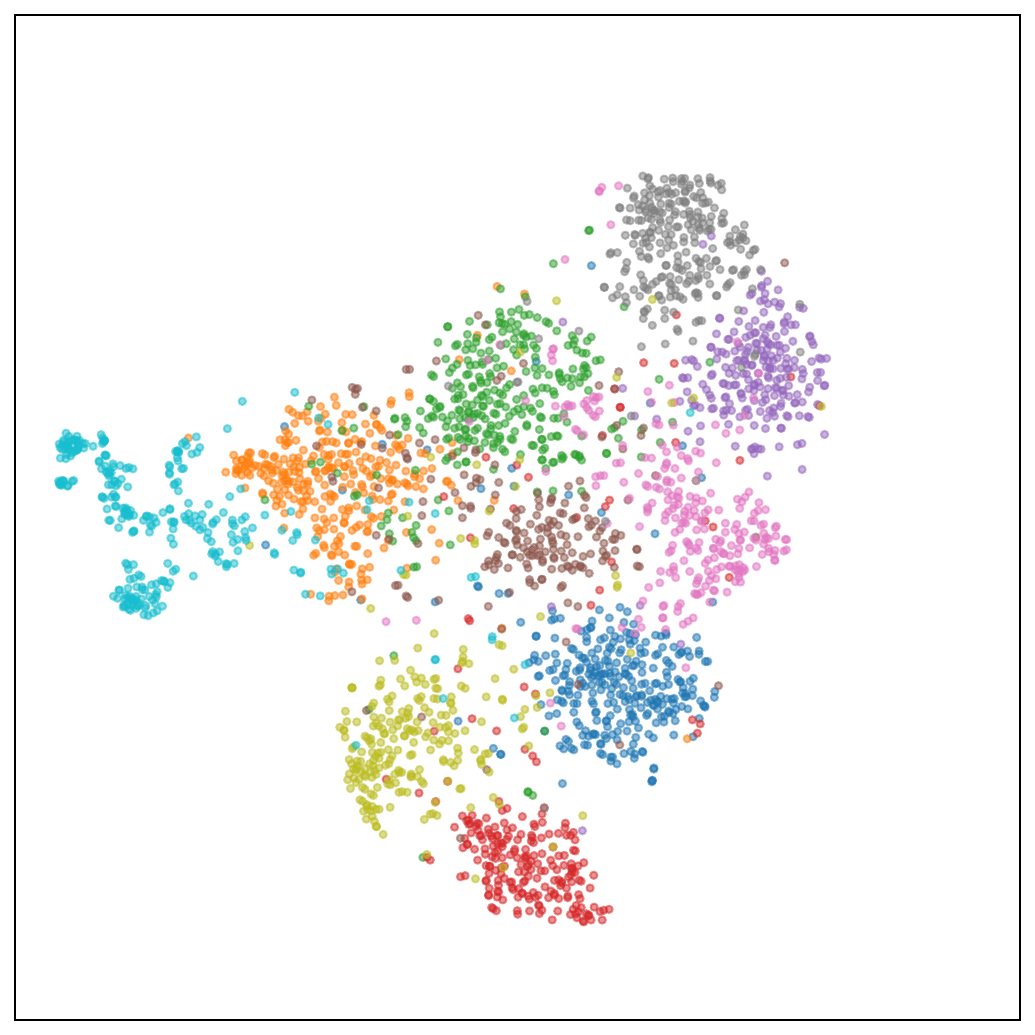}
        \caption{DOP}
    \end{subfigure}\hfill
    \begin{subfigure}{0.23\textwidth}
        \includegraphics[width=\linewidth]{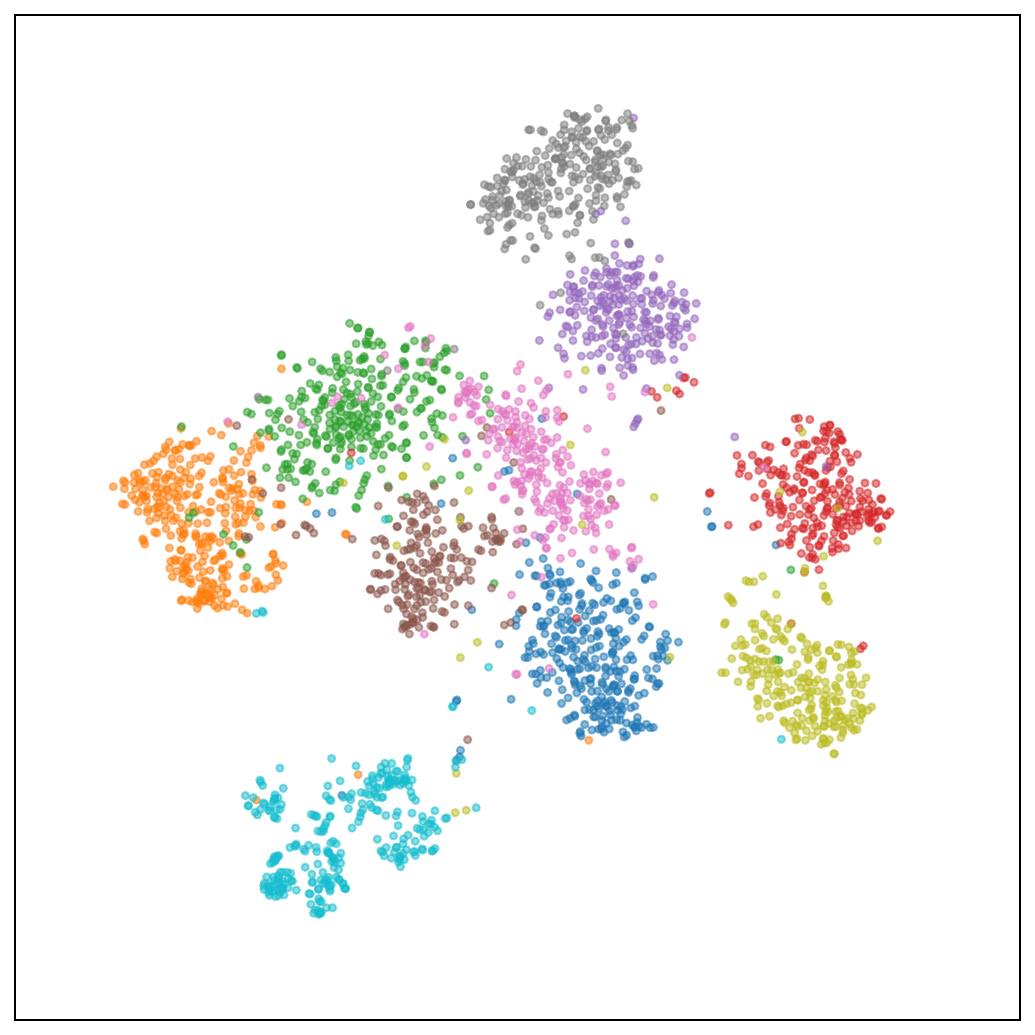}
        \caption{NUFILT}
    \end{subfigure}\hfill
    \begin{subfigure}{0.23\textwidth}
        \includegraphics[width=\linewidth]{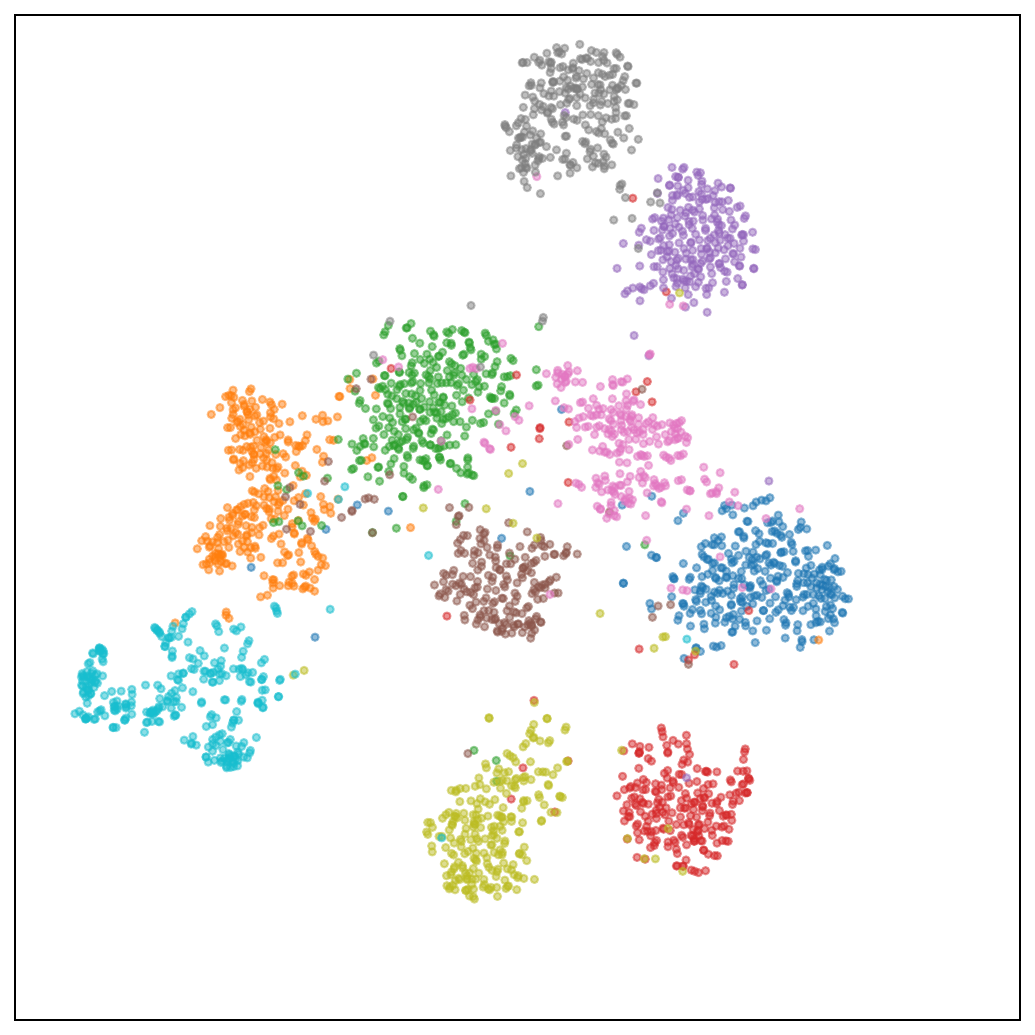}
        \caption{OURS}
    \end{subfigure}
    \caption{t-SNE visualization of the representation distributions of EuroSAT extracted from merged models using different methods.}
    \label{fig:tsne}
\end{figure}

\begin{table}
\centering
\caption{Effect of merging a repeated task. After merging 8 tasks with ViT-B/32, the task appearing fourth in each order (DTD for Order 1, SVHN for Order 2) is merged once more. The accuracy of the repeated task is underlined.}
\label{tab:repeated_tasks}
\begin{tabular}{c|cccccccc|c}
\toprule
& Cars & DTD & Euro & GTS & MNIST & RESIS & SVHN & SUN & AVG \\
\midrule
Order 1, original & 71.5 & \underline{63.6} & 95.0 & 91.6 & 98.7 & 84.2 & 89.8 & 67.4 & 82.7 \\
Order 1, repeated & 72.1 & \underline{68.8} & 94.7 & 92.9 & 98.8 & 84.9 & 91.2 & 67.8 & 83.9 \\
\midrule
Order 2, original & 70.5 & 63.2 & 95.3 & 92.8 & 98.7 & 83.0 & \underline{91.0} & 67.1 & 82.7 \\
Order 2, repeated & 68.7 & 61.1 & 91.0 & 90.2 & 98.4 & 80.9 & \underline{95.8} & 65.9 & 81.5 \\
\bottomrule
\end{tabular}
\end{table}

\subsection{Additional Qualitative Analysis}
\label{appendix:qual}
We provide t-SNE~\citep{tsne} visualizations of representations of the fourth merged EuroSAT task from a ViT-B/32 model merged over 20 tasks. As shown in Figure~\ref{fig:tsne}, prior methods fail to preserve structured representations, whereas our method maintains well-separated, class-consistent clusters. This qualitative difference aligns with the superior performance of the merged model and indicates that our approach better preserves task-specific representations during continual model merging.

\section{Limitations \& Broader Impact}
\label{appendix:broader}

\subsection{Limitations}
\paragraph{Scalability to longer task sequences}
Although our experiments include relatively long continual merging settings with up to 20 sequential tasks, we do not explicitly investigate regimes in which the accumulated task complexity substantially exceeds the representational capacity of the base model. As increasingly diverse tasks are merged into a fixed-capacity model, interference between task-specific representations may become unavoidable, potentially leading to performance saturation or degradation even when task balance is preserved. Understanding the relationship among model capacity, task diversity, and the maximum number of effectively mergeable tasks remains an important direction for future work.

\paragraph{Heterogeneous architectures}
Another limitation is that our experiments primarily focus on supervised fine-tuned checkpoints with relatively aligned architectures. The behavior of the proposed method under heterogeneous settings, such as models trained with different tokenizers, modalities, or optimization objectives, remains underexplored. Investigating robust merging across highly diverse foundation models could further broaden the applicability of continual model merging.

\subsection{Broader Impact}
This work aims to improve continual model merging while preserving task balance. Nevertheless, misuse of merged models could exacerbate risks such as membership inference attacks and safety leakage. Mitigating these risks requires that progress in model merging be paired with rigorous AI safety research and adherence to ethical guidelines, including the development of robust safeguards against information leakage and the incorporation of differential privacy techniques.



\end{document}